\documentclass[]{fairmeta} 

\usepackage{placeins}  

\usepackage{etoolbox}
\newcommand{\arxiv}[1]{\iftoggle{iclr}{}{#1}}
\newcommand{\iclr}[1]{\iftoggle{iclr}{#1}{}}
\newtoggle{iclr}
\global\toggletrue{iclr}
\global\togglefalse{iclr}

\usepackage{float}
\usepackage{booktabs}       %

\usepackage{graphics}
\usepackage{mathtools}

\usepackage{enumitem}

\usepackage{amsfonts}       %
\usepackage{amsmath}       %
\usepackage{amssymb}
\usepackage{amsthm}
\usepackage{pifont}%
\usepackage{diagbox}
\usepackage{multirow}

\makeatletter
\newcommand\notsotiny{\@setfontsize\notsotiny{6.31415}{7.1828}}
\makeatother

\newcommand{\Fig}[1]{Figure~\ref{#1}}  %
\newcommand{\fig}[1]{Fig.~\ref{#1}}    %

\newcommand{\tab}[1]{Table~\ref{#1}}

\renewcommand{\sec}[1]{Sec.~\ref{#1}} %

\usepackage{xspace}
\makeatletter
\DeclareRobustCommand\onedot{\futurelet\@let@token\@onedot}
\def\@onedot{\ifx\@let@token.\else.\null\fi\xspace}
\def\eg{e.g\onedot}
\def\ie{i.e\onedot}

\makeatother

\let\originalleft\left
\let\originalright\right
\renewcommand{\left}{\mathopen{}\mathclose\bgroup\originalleft}
\renewcommand{\right}{\aftergroup\egroup\originalright}

\newcommand{\g}[1]{%
  \ifthenelse{\equal{#1}{(}}
  {\left( }%
    { \ifthenelse{\equal{#1}{)}}
      { \right)}%
    { \ifthenelse{\equal{#1}{[}}
      {\left[}%
        { \ifthenelse{\equal{#1}{]}}
          { \right]}%
        {#1}}
    }
  }
}

\definecolor{ourblue}{rgb}{0.368,0.507,0.71}
\definecolor{ourorange}{rgb}{0.881,0.611,0.142}
\definecolor{ourgreen}{rgb}{0.56,0.692,0.195}
\definecolor{ourred}{rgb}{0.923,0.386,0.209}
\definecolor{ourviolet}{rgb}{0.528,0.471,0.701}
\definecolor{ourbrown}{rgb}{0.772,0.432,0.102}
\definecolor{ourlightblue}{rgb}{0.364,0.619,0.782}
\definecolor{ourdarkgreen}{rgb}{0.572,0.586,0.}
\definecolor{ourdarkblue}{RGB}{28,99,148}
\definecolor{ourdarkred}{RGB}{169,53,17}

\usepackage{etoolbox}
\makeatletter
\patchcmd{\NAT@test}{\else \NAT@nm}{\else \NAT@nmfmt{\NAT@nm}}{}{}
\DeclareRobustCommand\citeposs
  {\begingroup
   \let\NAT@nmfmt\NAT@posfmt%
   \NAT@swafalse\let\NAT@ctype\z@\NAT@partrue
   \@ifstar{\NAT@fulltrue\NAT@citetp}{\NAT@fullfalse\NAT@citetp}}

\let\NAT@orig@nmfmt\NAT@nmfmt
\def\NAT@posfmt#1{\NAT@orig@nmfmt{#1's}}
\makeatother

\usepackage{xr-hyper}

\makeatletter
\newcommand*{\addFileDependency}[1]{%
  \typeout{(#1)}
  \@addtofilelist{#1}
  \IfFileExists{#1}{}{\typeout{No file #1.}}
}
\makeatother

\arxiv{
\usepackage{mathtools}
\usepackage{amsmath} 
\usepackage{amsfonts}
\usepackage{bbm}
\usepackage[most]{tcolorbox}
\definecolor{midgrey}{RGB}{200, 200, 200}%
\usepackage[]{minted}
\usepackage{adjustbox}

\renewcommand{\epsilon}{\varepsilon}

\def\ddefloop#1{\ifx\ddefloop#1\else\ddef{#1}\expandafter\ddefloop\fi}
\def\ddef#1{\expandafter\def\csname bb#1\endcsname{\ensuremath{\mathbb{#1}}}}
\ddefloop ABCDEFGHIJKLMNOPQRSTUVWXYZ\ddefloop
\def\ddefloop#1{\ifx\ddefloop#1\else\ddef{#1}\expandafter\ddefloop\fi}
\def\ddef#1{\expandafter\def\csname b#1\endcsname{\ensuremath{\mathbf{#1}}}}
\ddefloop ABCDEFGHIJKLMNOPQRSTUVWXYZ\ddefloop
\def\ddef#1{\expandafter\def\csname sf#1\endcsname{\ensuremath{\mathsf{#1}}}}
\ddefloop ABCDEFGHIJKLMNOPQRSTUVWXYZ\ddefloop
\def\ddef#1{\expandafter\def\csname c#1\endcsname{\ensuremath{\mathcal{#1}}}}
\ddefloop ABCDEFGHIJKLMNOPQRSTUVWXYZ\ddefloop
\def\ddef#1{\expandafter\def\csname h#1\endcsname{\ensuremath{\widehat{#1}}}}
\ddefloop ABCDEFGHIJKLMNOPQRSTUVWXYZ\ddefloop
\def\ddef#1{\expandafter\def\csname hc#1\endcsname{\ensuremath{\widehat{\mathcal{#1}}}}}
\ddefloop ABCDEFGHIJKLMNOPQRSTUVWXYZ\ddefloop
\def\ddef#1{\expandafter\def\csname t#1\endcsname{\ensuremath{\widetilde{#1}}}}
\ddefloop ABCDEFGHIJKLMNOPQRSTUVWXYZ\ddefloop
\def\ddef#1{\expandafter\def\csname tc#1\endcsname{\ensuremath{\widetilde{\mathcal{#1}}}}}
\ddefloop ABCDEFGHIJKLMNOPQRSTUVWXYZ\ddefloop
\def\ddefloop#1{\ifx\ddefloop#1\else\ddef{#1}\expandafter\ddefloop\fi}
\def\ddef#1{\expandafter\def\csname scr#1\endcsname{\ensuremath{\mathscr{#1}}}}
\ddefloop ABCDEFGHIJKLMNOPQRSTUVWXYZ\ddefloop

}

\iclr{
\usepackage{iclr2026_conference, times}

\usepackage[utf8]{inputenc} 
\usepackage[T1]{fontenc}    
\usepackage{url}            
\usepackage{booktabs}       
\usepackage{amsfonts}       

\usepackage{amsmath} 
\usepackage{mathtools}
\usepackage{nicefrac}       
\usepackage{microtype}      

\usepackage{tocloft}            
\usepackage{subcaption}

\usepackage{enumitem}

\usepackage{breakcites}

\newtoggle{draft}
\togglefalse{draft}

\usepackage{mathrsfs}

\usepackage{algorithm}
\usepackage{verbatim}
\usepackage[noend]{algpseudocode}

\usepackage{multicol}

\usepackage{colortbl}

\usepackage{setspace}

\usepackage{transparent}

\usepackage{inconsolata}
\iclr{\usepackage[scaled=.90]{helvet}}
\usepackage{xspace}

\usepackage{pifont}

\usepackage{hyperref}
\newcommand{\alghyperref}[1]{\hyperref[#1]{Alg.~\ref*{#1}}}
}

\usepackage[suppress]{color-edits}
\addauthor{ys}{MidnightBlue}

\usepackage[most]{tcolorbox}
\usepackage{xcolor}

\definecolor{paperblue}{HTML}{1877F2}

\newtcolorbox{takeawaybox}{
  colback=paperblue!6,      
  colframe=paperblue!65,    
  title=\textbf{Takeaway},
  boxrule=0.3pt,
  arc=2pt,
  left=6pt,
  right=6pt,
  top=4pt,
  bottom=4pt,
  fonttitle=\bfseries,
}

\usepackage{listings}
\usepackage{mdframed}

\tcbuselibrary{breakable, skins}

\newcommand{\sectionlabel}[2]{%
  \noindent\colorbox{#1!20}{\parbox{\dimexpr\linewidth-2\fboxsep}{\small\textbf{#2}}}%
  \vspace{2pt}%
}

\newmdenv[
  linecolor=paperblue!65,
  linewidth=2pt,
  topline=false,
  bottomline=false,
  rightline=false,
  innerleftmargin=10pt,
  innertopmargin=6pt,
  innerbottommargin=6pt,
  innerrightmargin=0pt,
  backgroundcolor=paperblue!3,
  skipabove=6pt,
  skipbelow=6pt
]{turntwoPromptFrame}

\newmdenv[
  linecolor=orange!65,
  linewidth=2pt,
  topline=false,
  bottomline=false,
  rightline=false,
  innerleftmargin=10pt,
  innertopmargin=6pt,
  innerbottommargin=6pt,
  innerrightmargin=0pt,
  backgroundcolor=orange!3,
  skipabove=6pt,
  skipbelow=6pt
]{turntwoAnswerFrame}

\newtcolorbox{reasoningbox}[1]{
  breakable,
  enhanced,
  colback=white,
  colframe=orange!30,
  title=#1,
  boxrule=0.3pt,
  arc=2pt,
  left=4pt, right=4pt, top=3pt, bottom=3pt,
}

\title{A Deep Dive into Scaling RL for Code Generation with Synthetic Data and Curricula}

\arxiv{
\subtitle{}

\usepackage{xspace}

\author[1,2,*]{Cansu Sancaktar}
\author[3]{David Zhang}
\author[3]{Gabriel Synnaeve}
\author[3]{Taco Cohen}

\affiliation[1]{University of Tübingen}
\affiliation[2]{Max Planck Institute for Intelligent Systems}
\affiliation[3]{Meta FAIR}

\contribution[*]{Work done during an internship at Meta}

\date{\today}
\correspondence{Cansu Sancaktar at \email{cansu.sancaktar@tuebingen.mpg.de}}
\abstract{
Reinforcement learning (RL) has emerged as a powerful paradigm for improving large language models beyond supervised fine-tuning, yet sustaining performance gains at scale remains an open challenge, as data diversity and structure, rather than volume alone, become the limiting factor. We address this by introducing a scalable multi-turn synthetic data generation pipeline in which a teacher model iteratively refines problems based on in-context student performance summaries, producing structured difficulty progressions without any teacher fine-tuning. Compared to single-turn generation, this multi-turn approach substantially improves the yield of valid synthetic problems and naturally produces stepping stones, i.e. easier and harder variants of the same core task, that support curriculum-based training. We systematically study how task difficulty, curriculum scheduling, and environment diversity interact during RL training across the Llama3.1-8B Instruct and Qwen3-8B Base model families, with additional scaling experiments on Qwen2.5-32B. Our results show that synthetic augmentation consistently improves in-domain code and in most cases out-of-domain math performance, and we provide empirical insights into how curriculum design and data diversity jointly shape RL training dynamics.
}

}

\begin{document}

\maketitle

\iclr{
\begin{abstract}
\end{abstract}
}

\begin{figure}[h!]
    \centering
    \includegraphics[width=1.0\linewidth]{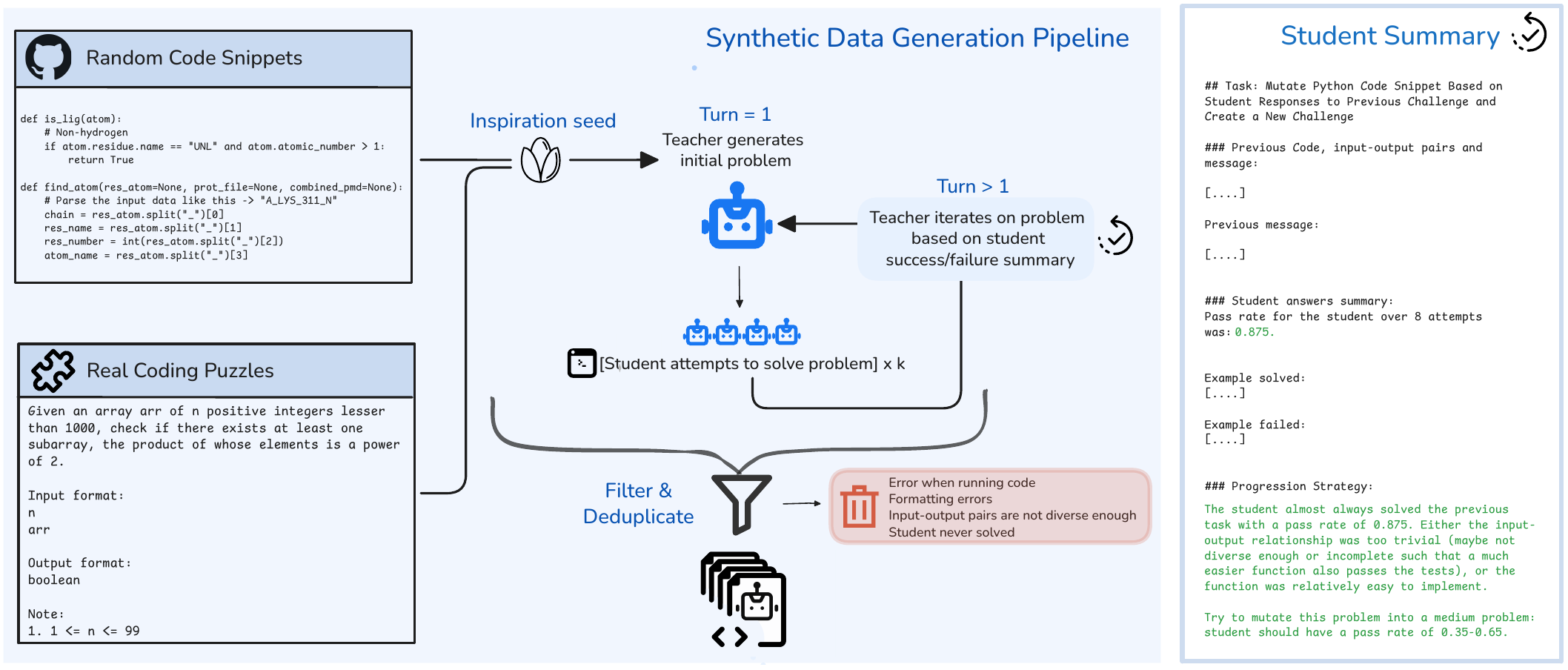}
    \caption{\textbf{Overview of the multi-turn synthetic data pipeline.} A seed snippet, sampled from random code or real coding puzzles, serves as inspiration for the teacher. In the first turn, the teacher generates an initial problem according to the current environment’s rules, and the student attempts to solve it multiple times. In later turns, the teacher receives a summary of the student’s performance (pass rate and representative solutions) and adapts the problem accordingly. Invalid or redundant generations are filtered and deduplicated before inclusion in the dataset.} 
    \label{fig:overview}
\end{figure}

\section{Introduction}
Reinforcement learning (RL) has become a central paradigm for improving large language models (LLMs) beyond supervised fine-tuning. Yet as RL scales, a key challenge emerges: how can we sustain performance gains, and where should new training data come from? Addressing this question requires understanding what defines an effective RL task and environment.
Recent works explore asymmetric play in the LLM post-training era, where a teacher dynamically generates data tailored to the student model’s learning progress, aiming for open-ended RL at scale. However, effectively training such a teacher remains elusive. As shown in \citet{zhao2025absolute}, training the teacher does not result in substantial improvements over training only the student. Although there are positive gains from dynamically generating synthetic tasks, we are still bottlenecked by a limited teacher such that initial performance gains plateau.
\paragraph{\textbf{RL with real data is not easy to scale.}} One fundamental challenge with RL post-training is that naively increasing the number of problems does not guarantee further performance gains, as results are confounded by RL dynamics as well as the difficulty distribution of the problem set at hand. Typically, each problem set has a mix of easy and medium-level problems, accompanied by very hard problems. The latter split poses a hard exploration challenge that cannot be solved by the initial model, such that the computation spent on these samples early on in training hurts efficiency. However, on the other end of the spectrum, the very easy questions give early gradient updates, but dominate training while the model's entropy collapses. 

A common strategy proposed in the literature is curriculum sampling, where the model is first trained on easier problems, gradually shifting towards the hard ones \citep{team2025kimi}. However, in the case of real data, there is often no guarantee that the hard problems build upon the easy ones in a meaningful way. Moreover, since model entropy naturally decreases during RL training \citep{cui2025entropy}, spending too much of the exploration budget on easy problems can reduce the model’s capacity to learn from more challenging ones later. These factors limit the effectiveness of naive curriculum-based sampling in practice.

Taking inspiration from the idea of stepping stones in goal-based exploration~\citep{forestier2022goal}, we incorporate easier variants of hard problems to help alleviate the hard exploration challenge.
We propose a multi-turn approach to generate synthetic data (\fig{fig:overview}). Similar to Magicoder \citep{wei2023magicoder}, we make use of actual code snippets to ground the problem generation and ensure diversity and coverage. 
Unlike standard single-turn generation, where multiple problem instances are independently sampled from a seed snippet and later filtered for solvability, formatting, and diversity, our method introduces an additional iteration axis. The teacher revisits and mutates the initially generated problem over multiple turns, adjusting its difficulty at each step. This process produces structured variants of the same core task, yielding easier and harder counterparts that serve as targeted augmentations. Crucially, this adaptation happens purely through in-context learning, requiring no gradient updates to the teacher.
Multi-turn data generation provides a practical pipeline for producing high-quality synthetic data at scale, reducing invalid generations and naturally introducing problem variants of varying difficulty.
Augmenting real code-contest questions with such synthetic data results in consistent performance gains on both in-domain (code) and out-of-domain (math) benchmarks.

To better understand the individual factors contributing to RL improvements, we systematically isolate the effects of different design choices. In particular, we address the following research questions:
\begin{itemize}
\item Do \textit{stepping stones}, \ie intermediate problems of varying difficulty, improve downstream generalization?
\item How does the difficulty level of RL problems influence training dynamics and convergence?
\item To what extent does the choice of curriculum schedule matter?
\item Given a limited compute budget, is it more effective to scale the number of problems within a single environment, or to scale across multiple environments; and can environment diversity serve as an independent axis of improvement?
\end{itemize}

Our main contributions are a scalable multi-turn framework for synthetic data creation and a systematic experimental study of how task difficulty, curriculum design, and environment diversity affect RL for code generation. Through extensive experiments across model families (Llama3.1-8B Instruct, Qwen3-8B Base, and Qwen2.5-32B Base), we analyze convergence and generalization under different data generation and training setups.
  
\section{Related Work}

\paragraph{Synthetic Data.}
Synthetic data generation has become a powerful tool both in supervised fine-tuning (SFT) and reinforcement learning (RL) for large language models (LLMs). Magicoder \citep{wei2023magicoder} demonstrates that large-scale synthetic code data can substantially enhance code generation capabilities of a model in an SFT setting, motivating the use of automatically generated problem instances for RL. Similarly, \citet{havrilla2025sparq} present SPARQ, which generates over 20 million synthetic math problems using quality–diversity algorithms and studies how problem difficulty and diversity affect generalization in fine-tuning. \citet{jiang2025generative} explore generative data refinement, transforming noisy real-data corpora into higher-utility training data for model adaptation.

At the RL stage, \citet{guo2025synthetic} propose a framework for RL using only synthetic question-answer pairs derived from task definitions and retrieved documents, reporting large improvements across reasoning benchmarks. \citet{liang2025sws} introduce \emph{SwS}, which identifies model weakness areas and synthesizes targeted problems to improve reasoning performance. \citet{goldie2025synthetic} further address multi-step reasoning and tool-use tasks via synthetic data generation in a multi-step RL pipeline. Together, these works highlight the increasing role of synthetic data in scaling adaptation and reasoning for LLMs. Our approach builds on this line of work by combining structured RL environments with teacher–student generation and curriculum control to optimize both in-domain performance and out-of-domain generalization.

\paragraph{Asymmetric Self-Play.}
Asymmetric self-play in LLMs has been explored across several domains, including coding and verification \citep{zhao2025absolute, lin2025learning}, tool use \citep{zhou2025self}, alignment \citep{ye2025evolving}, and theorem proving \citep{dong2025stp, poesia2024learning}. These works leverage the interaction between a generator (teacher) and a solver (student) to create adaptive task distributions, an idea closely related to our teacher–student setup for multi-turn data generation.

\paragraph{Curriculum Learning.}
Curriculum learning remains an active area of research in RL for LLM post-training. 
\citet{wang2025dump} propose a distribution-level curriculum learning framework, where the sampling probabilities are dynamically adjusted, prioritizing either distributions with high average advantage (exploitation) or low sample count (exploration).
\citet{bae2025online} introduce adaptive sampling strategies that remove overly easy or hard examples to focus on intermediate difficulty. In the context of alignment, \citet{pattnaik2024curry} introduce curricula that gradually increase task difficulty, while \citet{team2025kimi} employ manually defined heuristic curriculum stages, starting with easy problems and gradually progressing to harder ones. They also use a prioritized sampling strategy, where problems with lower success rates receive higher sampling probabilities.
\citet{xie2025logic} study the effectiveness of reasoning with RL on synthetic logic puzzles, showing that under a fixed data budget, well-designed curricula consistently outperform random shuffling. Finally, reverse curricula have been shown to improve exploration, where the start state of reasoning progressively slides from a demonstration's end to its beginning \citep{xi2024training}.

\section{Synthetic Data Generation Pipeline}

\subsection{Environments}
Each environment is framed as a teacher–student interaction: the teacher generates task instances, and the student attempts to solve them.
The RL environments used throughout this work are: \texttt{Induction}, \texttt{Abduction}, \texttt{Deduction} (inspired by \citet{zhao2025absolute}) and \texttt{Fuzzing}. 

\paragraph{Induction} Program synthesis environment. This can be seen as a variant of classical code puzzle environments used for RL post-training.  
\paragraph{Abduction} Input prediction environment. Given a function \texttt{f()} and an output $o$, the task is to predict the input $i$ that produced $o$.  
\paragraph{Deduction} Output prediction environment. Given a function \texttt{f()} and an input $i$, the task is to predict the output $o$ obtained upon execution.  
\paragraph{Fuzzing} Fuzzing environment inspired by property-based testing. Given \texttt{f()}, \texttt{pre\_test\_f()}, and \texttt{test\_f()}, the task is to find an input such that \texttt{test\_f()} fails while \texttt{pre\_test\_f()} passes. The function \texttt{pre\_test\_f()} acts as a type checker, \eg verifying correct input types to ensure that \texttt{test\_f()} fails for non-trivial reasons.  

The corresponding teacher–student specifications and reward definitions for each environment are summarized in \tab{tab:envs}.

\begin{table}[h!]
\centering
\caption{\textbf{Overview of RL environments.} The teacher column specifies how each task instance is generated, while the student column describes the corresponding RL problem to be solved. The final column shows the reward function used for evaluation. Here, $\hat{\cdot}$ denotes the student’s prediction, and $\mathbb{I}[\cdot]$ is the indicator function.}
\label{tab:envs}
\begin{adjustbox}{width=\textwidth,center}
\begin{tabular}{p{2.5cm} p{5.5cm} p{5.5cm} p{5.5cm}}
\toprule
\textbf{Environment} & \textbf{Teacher: Generator} & \textbf{Student: Solver}  & \textbf{Reward}\\
\midrule
\texttt{Induction} & Generates function \texttt{f()}, a natural language \texttt{message}, and $k$ input cases ${i_1, ..., i_k}$. The corresponding gold outputs ${o_1, ..., o_k}$ are computed by execution: $o_k = \mathtt{f}(i_k)$. & Given $k' < k$ input-output pairs ${\{i_1,o_1\} ..., \{i_k', o_k'\}}$ (the remaining $k - k'$ are held out as private test cases) and the \texttt{message}, student synthesizes \texttt{f()}. & $r = \mathbb{I}\!\left[\forall k,\ \hat{f}(i_k) = o_k\right]$\\ \\
\texttt{Abduction} & Generates function \texttt{f()} and one input case $i$. The gold output is computed as $o=\mathtt{f}(i)$. & Given \texttt{f()} and the output \texttt{o}, infers the input \texttt{i} that produced the given output.  & $r = \mathbb{I}[f(\hat{i}) = o]$\\ \\
\texttt{Deduction} & Generates function \texttt{f()} and one input case $i$. The gold output is computed as $o=\mathtt{f}(i)$. & Given \texttt{f()} and the input \texttt{i}, infers the output \texttt{o} to be obtained when executing \texttt{f(i)}. & $r = \mathbb{I}[\hat{o} = o]$\\ \\
\texttt{Fuzzing} & Generates a function \texttt{f()}, which contains a subtle bug to be exploited, and tests \texttt{pre\_test\_f()} and \texttt{test\_f()}. & Given \texttt{f()}, \texttt{pre\_test\_f()} and \texttt{test\_f()}, finds an input test case such that \texttt{test\_f()} fails while passing \texttt{pre\_test\_f()}. & $r = \mathbb{I}[\texttt{pre\_test\_f}(\hat{i}) \land \neg\,\texttt{test\_f}(\hat{i})]$ \\ \\
\bottomrule
\end{tabular}
\end{adjustbox}
\end{table}

\subsection{Inspiration seeds} \label{sec:insp}

We use seed data as inspiration for synthetic data generation to ensure diversity and grounding. We use two different classes of seed data: (1) we extract successful solutions to real coding problems solved during an initial RL training run, and (2) random open source code snippets collected from the wild.

For (2), similar to \citet{wei2023magicoder}, we adopt \texttt{starcoderdata} as our seed corpus. In order to generate a synthetic question, we randomly select 25--50 consecutive lines to be our seed snippet for inspiration, and the teacher model is asked to try to incorporate the core logic of this example snippet for data generation.

\subsection{Generating Multi-Turn Synthetic Data} \label{sec:multi-turn}
\begin{figure}[h!]
    \centering
    \includegraphics[width=1.03\linewidth]{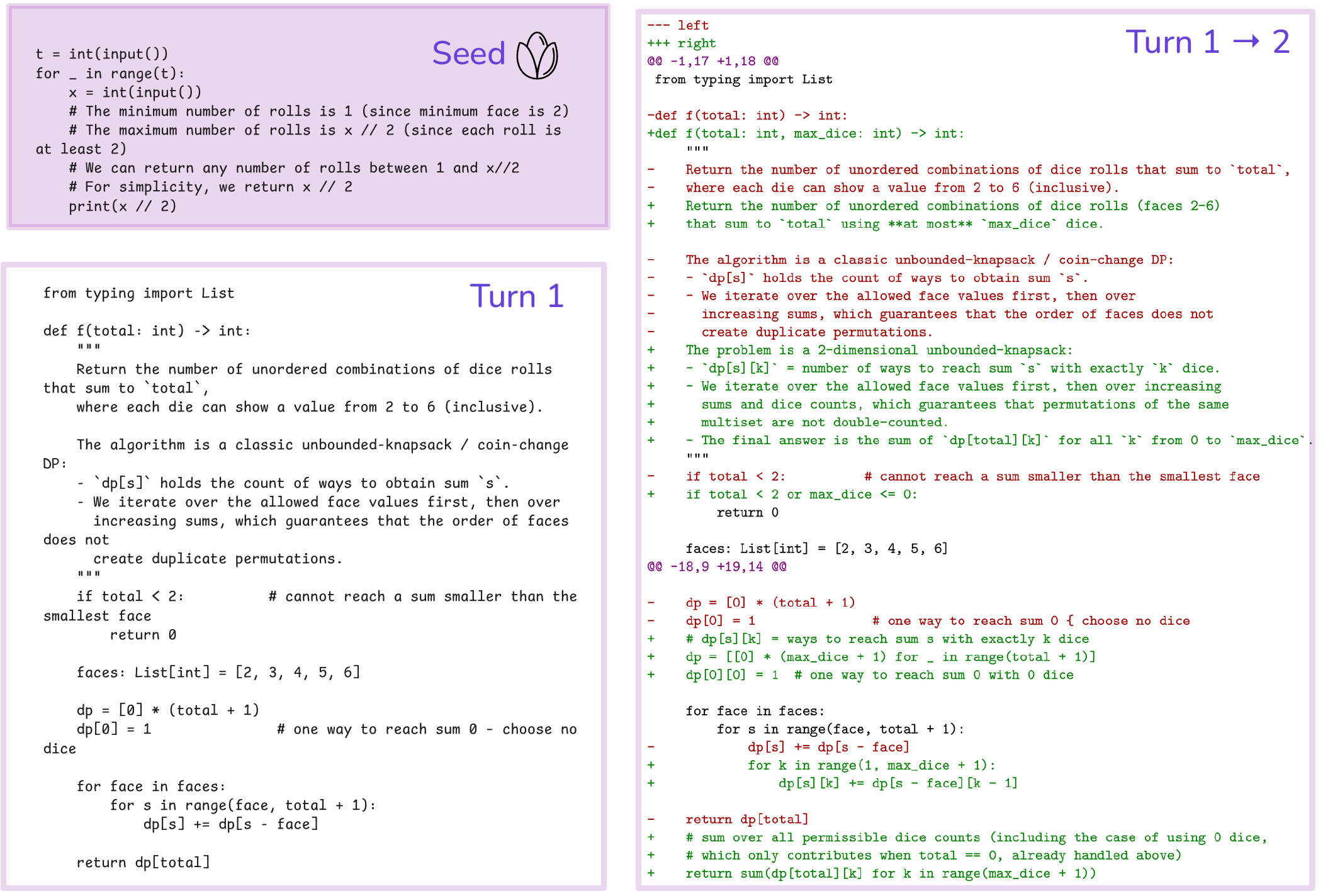}
    \caption{\textbf{Example of multi-turn data generation.} The top-left panel shows the seed snippet provided to the teacher, taken from a real coding puzzle. In turn 1, the teacher generates a puzzle with a student pass rate of 0.875 ($M=8$). In turn 2, after observing the student’s performance, the teacher produces a harder variant with a pass rate of 0.25.
}
    \label{fig:example}
\end{figure}

We perform multi-turn data generation, where the teacher iteratively adapts task difficulty based on the student's performance (\fig{fig:example}).

In the first turn, given a seed snippet (see \sec{sec:insp}) and a task description, the teacher is prompted to generate a valid problem instance following the rules of the current RL environment. The resulting problem $t_1$ is then presented to the student, who attempts to solve it $M$ times (\eg $M=32$).\footnote{In practice, we use $M=32$. We also experimented with $M=8$, but found that student pass rate estimates become noisy for small $M$.}

In the second turn, the student's attempts from turn 1 are summarized as the pass rate $p=\frac{1}{M}\sum_{m=1}^{M} r_m$, together with representative examples of both successful and failed solutions (if available). The teacher then receives the original question and this performance summary, and is prompted to adapt the problem accordingly for the next iteration. The adaptation follows a progression strategy that adjusts task difficulty based on the observed pass rate (\eg increasing complexity if $p>0.65$, or decreasing complexity if $p=0$). Subsequent turns follow the same iterative structure, with each turn conditioned on the immediately preceding question and student summary rather than the full history (in practice, we use 6 turns per seed). A full example of this in-context adaptation, including the teacher prompt and reasoning, is provided in \sec{sec:qualitative-example}.

Currently, the data generation pipeline is decoupled from the RL runs and the actual student. During multi-turn generation, the same model used as teacher takes on the role of the student. In our experiments, we use \texttt{GPT-OSS 120B} high reasoning mode as our teacher and the same model is used in low reasoning mode as the student. We expect further performance gains from having the teacher in the loop and learning from the mistakes of the student on-the-go, resulting in more effective augmentations. 

\subsection{Reinforcement Learning with Verifiable Rewards} \label{sec:rl}
We employ an asynchronous variant of Group Relative Policy Optimization (GRPO) \citep{shao2024deepseekmath} as our reinforcement learning algorithm. GRPO replaces the value function with a Monte Carlo–based estimation, sampling $G$ outputs from the model and computing the advantage of each $o_i$ ($i \in \{1,\dots,G\}$) by normalizing its reward relative to others within the group. In our implementation, we omit both the standard-deviation normalization in the advantage computation and the KL-regularization term (see \sec{sec:rl-hyper} for more details).

\section{Results}
We train Llama3.1-8B Instruct~\citep{grattafiori2024llama}, Qwen3-8B Base (non-thinking mode)~\citep{yang2025qwen3}, and Qwen2.5-32B Base~\citep{qwen2.5} with reinforcement learning on code tasks without explicit reasoning traces, and evaluate both in-domain (code) and out-of-domain (math) performance to analyze performance across model families and scales.
For code, we use \texttt{LiveCodeBench} (LCB) \citep{jain2024livecodebench}, aggregating queries between 08.2024 and 05.2025 (454 problems), as well as the \texttt{LCBv5} splits (\texttt{easy}, \texttt{medium}, \texttt{hard}; 05.2023–01.2025).  
For math, we evaluate on the \texttt{Math500} \citep{math500} and \texttt{AIME2024} benchmarks.
\subsection{Scaling and Augmenting RL with Synthetic Data} \label{sec:results_synthetic}
\begin{figure}[h]
    \centering
    \includegraphics[width=1.0\textwidth]{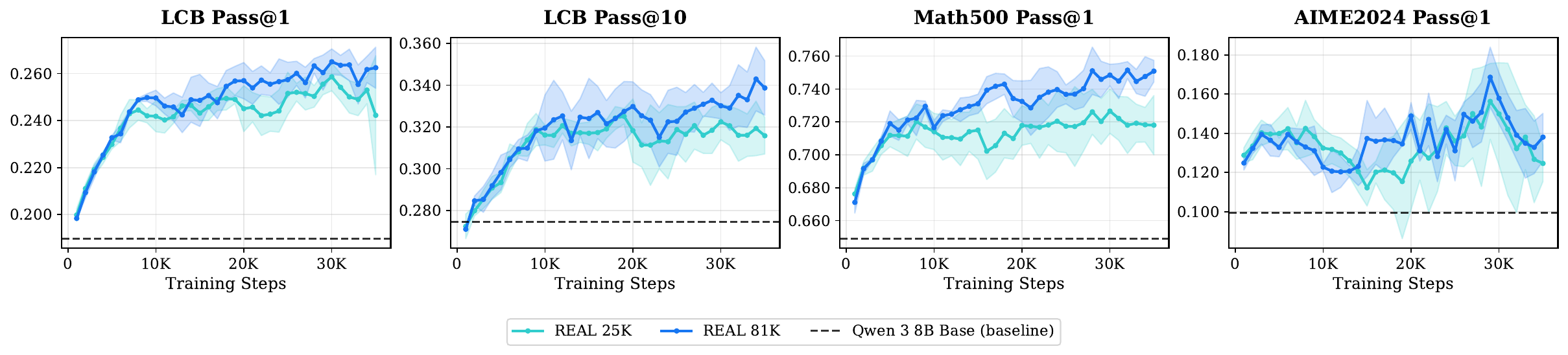}
\caption{\textbf{Scaling with real data in Qwen3-8B Base.}
We compare RL training on 25K and 81K real coding problems using GRPO (3 seeds).
Performance is tracked on in-domain (\texttt{LCB}) and out-of-domain (\texttt{Math500} and \texttt{AIME2024}) benchmarks throughout training.
Performance gains plateau early, indicating limited benefit from scaling real data alone.}

    \label{fig:qwen-real}
\end{figure}

We begin by examining whether scaling reinforcement learning with additional real coding problems alone leads to sustained performance gains. Figure~\ref{fig:qwen-real} compares RL training on 25K and 81K real coding problems in Qwen3-8B Base.

During RL training, performance improvements plateau early as policy entropy decreases. Increasing the number of RL problems does not yield proportional gains (Figure~\ref{fig:qwen-real}), suggesting that data diversity or structure, rather than volume alone, becomes the limiting factor.

\subsubsection{Does augmenting real coding problems with synthetic problems help boost performance?}
\begin{figure}[!h]
    \centering
    \includegraphics[width=.92\textwidth]{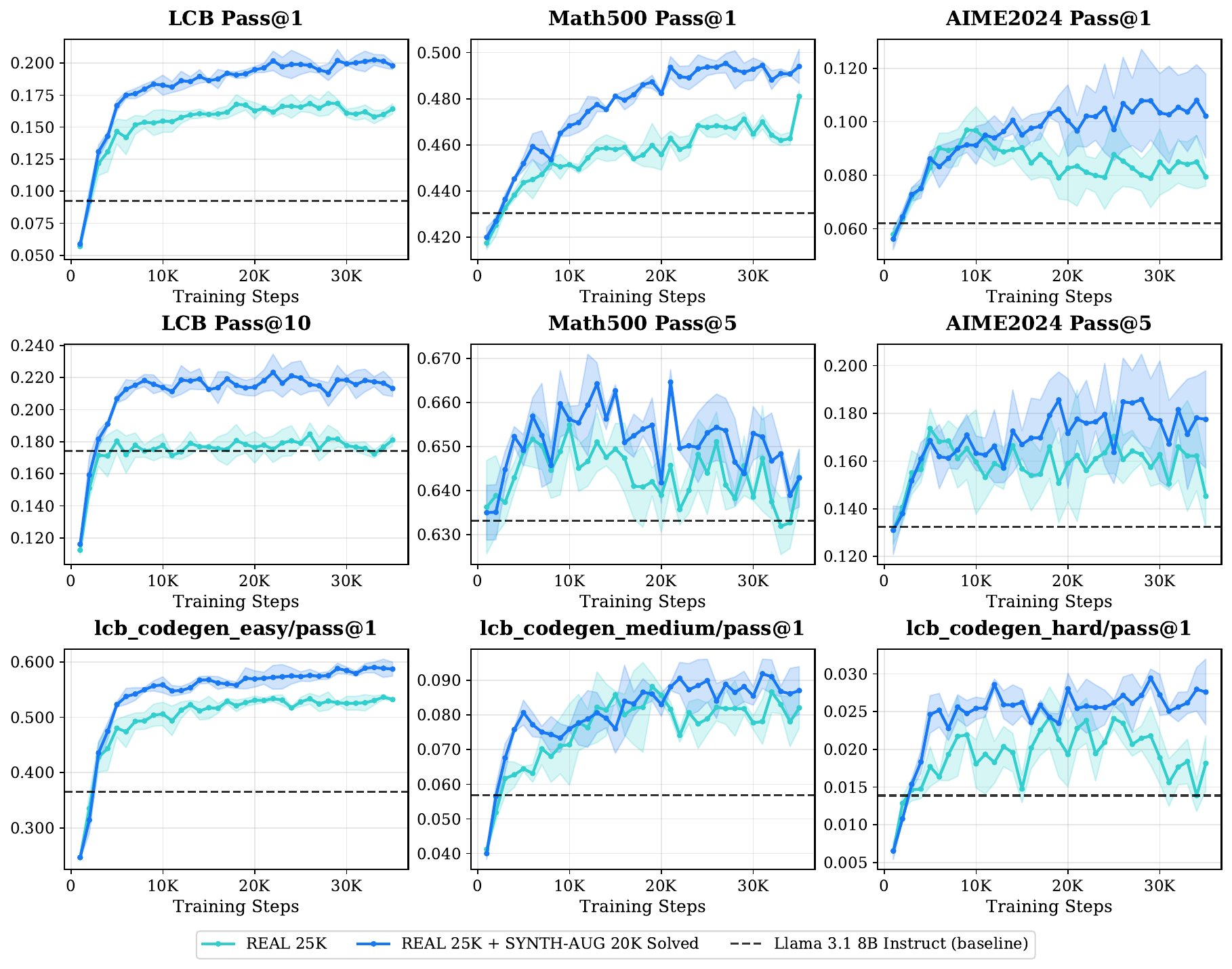}
    \vspace{-.8em}
\caption{\textbf{Synthetic data augmentations in Llama3.1-8B Instruct.}
RL training on 25K real code-contest problems (baseline) versus 25K real plus 20K synthetic problem augmentations seeded from solved real questions (3 seeds).
Synthetic augmentation improves performance across both in-domain (code: \texttt{LCB}) and out-of-domain (math: \texttt{Math500}, \texttt{AIME2024}) benchmarks.}
    \label{fig:llama-synth-aug}
\end{figure}

\begin{figure}[!h]
    \centering
    \includegraphics[width=1.0\textwidth]{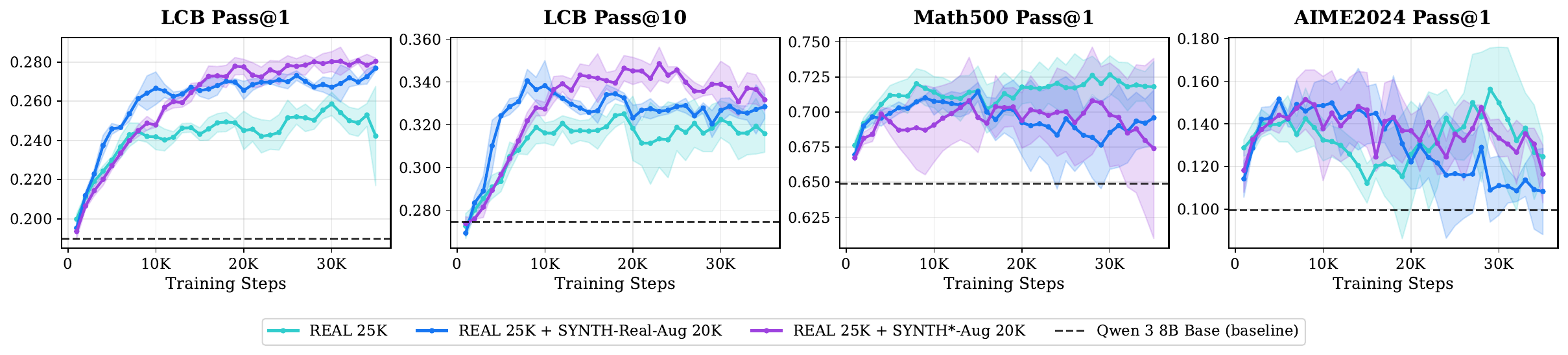}
    \vspace{-2em}
\caption{\textbf{Synthetic data augmentations in Qwen3-8B Base.}
RL training on real code-contest problems (baseline) versus with synthetic problem augmentation (3 seeds). Synthetic data are seeded either with answers to real questions (\texttt{SYNTH-Real-Aug}) or with random code snippets from \texttt{starcoderdata} (\texttt{SYNTH*-Aug}). Performance improves primarily on in-domain (code: \texttt{LCB}), while out-of-domain (math: \texttt{Math500}, \texttt{AIME2024}) benchmark performance remains comparable or slightly lower.}
    \label{fig:qwen-synth-aug}
\end{figure}

\begin{figure}[h]
    \centering
    \includegraphics[width=1.0\textwidth]{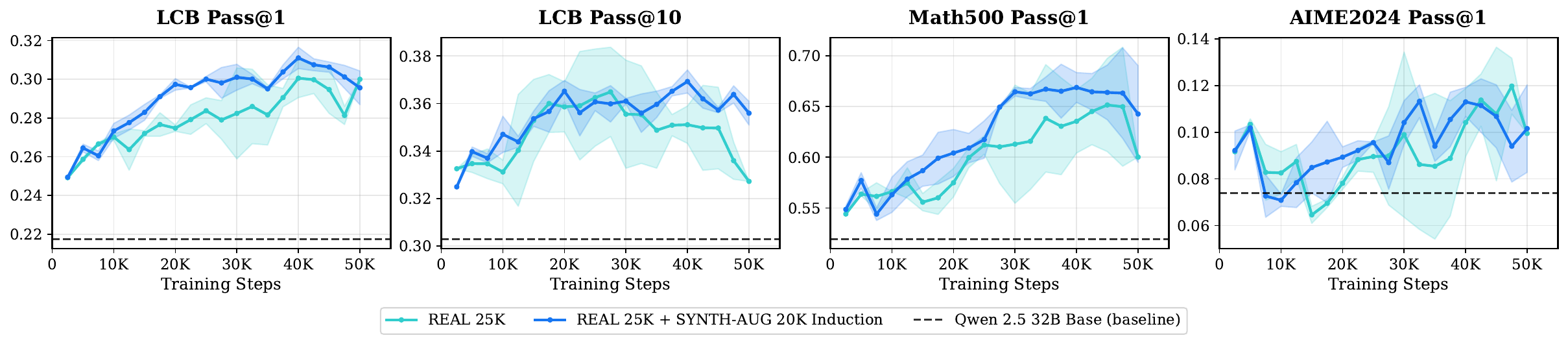}
    \caption{\textbf{Synthetic data augmentations in Qwen2.5-32B Base.}
RL training on 25K real code-contest problems (baseline) versus 25K real plus 20K synthetic problem augmentations seeded from solved real questions (2 seeds).
Synthetic augmentation shows improved performance trends for both in-domain (code: \texttt{LCB}) and out-of-domain (math: \texttt{Math500}, \texttt{AIME2024}) benchmarks.}
\label{fig:qwen32-synth-aug}
\end{figure}

We show that augmenting real coding questions with synthetic problems leads to faster and more stable convergence in-domain (code) for Llama3.1-8B Instruct (\fig{fig:llama-synth-aug}), Qwen3-8B Base (\fig{fig:qwen-synth-aug}), and Qwen2.5-32B Base (\fig{fig:qwen32-synth-aug}), and improves out-of-domain (math) performance for Llama3.1-8B Instruct and Qwen2.5-32B. 
For RL runs with augmented data, the total number of training problems is increased by 20K, while keeping the overall training budget fixed. Consequently, each problem is encountered fewer times within the same number of RL steps shown in the plots. 
Notably, synthetic augmentation also outperforms the larger 81K real-data baseline on most in-domain \texttt{LCB} metrics, underscoring the efficiency of diverse synthetic additions even under a fixed compute budget (see \fig{fig:qwen-synth-aug-and-scale-full}).

To test whether the choice of seed affects these gains, we perform an ablation comparing synthetic problems seeded with answers to real questions versus random code snippets from \texttt{starcoderdata} (see \fig{fig:qwen-synth-aug}). We match dataset difficulty across conditions to ensure that observed trends are not driven by easier questions.
Training on synthetic problems seeded with \texttt{starcoderdata} increases data diversity compared to augmenting real problems with their own variations. Consequently, we observe higher pass@1 and pass@10 in-domain, indicating that greater diversity boosts performance and that random-code seeding can be competitive with real data for synthetic augmentation.
Next, we analyze performance when training exclusively on synthetic problems.
\subsubsection{Are synthetic RL problems alone sufficient, and does seeding matter?}

\Fig{fig:qwen-synth-aug-real-full} and \Fig{fig:qwen-synth-aug-starcoder-full} show that RL training on fully synthetic problems can match the \texttt{LCB} scores achieved with real data. 
As seen in \fig{fig:qwen-synth-aug}, the difference between using real code-contest questions and random \texttt{starcoderdata} snippets as seeds for synthetic data generation is minimal. In the in-domain code setting (\texttt{LCB}), we observe slightly faster convergence when synthetic problems are seeded with real data. 
For all experiments, we match the difficulty histograms (based on student pass rates) between the \texttt{starcoderdata}- and real-seeded datasets to ensure a fair comparison.

Closer inspection reveals that most of the \texttt{LCB} gains are correlated with performance on the \texttt{LCBv5-easy} split, while performance on \texttt{medium} decreases. We hypothesize that this stems from the distribution of generated question difficulties: during filtering, we include all questions with pass rates between 0.01 and 0.97 without explicitly flattening the difficulty distribution. In practice, this results in a dataset dominated by easier questions, leading to overfitting on simpler tasks. 
We further analyze the effects of difficulty filtering and curriculum strategies in the following sections.

\begin{figure}[H]
    \centering
    \includegraphics[width=1.0\textwidth]{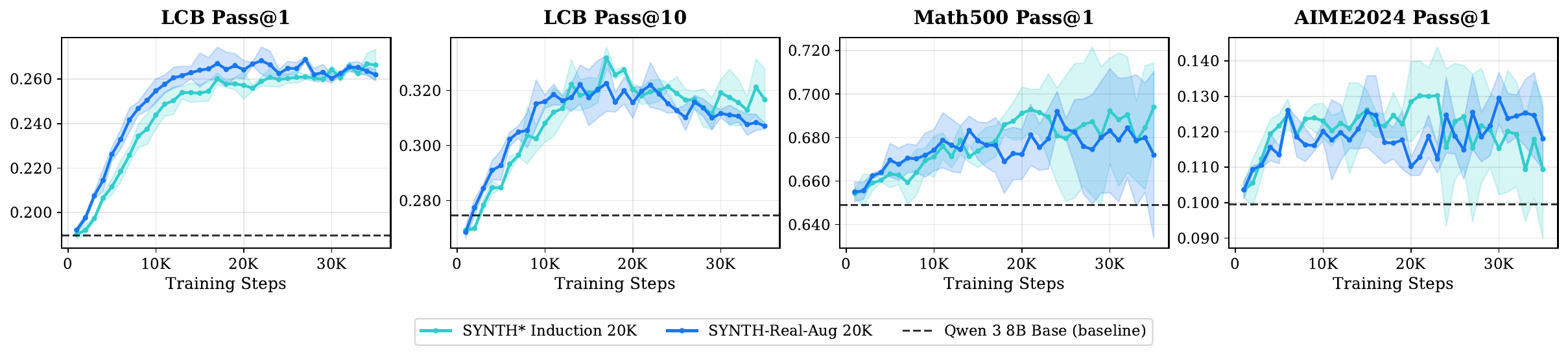}
\caption{\textbf{RL with pure synthetic data in Qwen3-8B Base.}
Instead of augmenting the 25K real code-contest questions, RL is performed solely on synthetic \texttt{induction} problems generated with two seeding strategies: 
(1) real questions as inspiration seeds (\texttt{SYNTH-Real-Aug}), and 
(2) random code snippets from \texttt{starcoderdata} (\texttt{SYNTH*-Aug}). Results shown for 3 seeds. 
Performance is comparable across both setups, with slightly faster in-domain convergence (\texttt{LCB}) when seeding with real data.}
    \label{fig:qwen-synth-only}
\end{figure}

\begin{takeawaybox}
\begin{itemize}
    \item Augmenting real code-contest questions with synthetic RL problems from a custom \texttt{induction} environment helps obtain better and faster convergence in-domain across model families, while also showing gains out-of-domain on Math benchmarks with Llama3.1-8B Instruct and Qwen2.5-32B Base.
    \item Diversifying inspiration snippets and using \texttt{starcoderdata} instead of the real data itself for augmentation shows improved trends in \texttt{LCB}.
    \item Synthetic RL problems alone are competitive; however, the difficulty distribution of the generated data might need explicit curation to avoid overfitting on \texttt{easy} problems.
    \item Seeding with real code-contest questions instead of \texttt{starcoderdata} provides slightly faster convergence on \texttt{LCB}.
\end{itemize}
\end{takeawaybox}

\subsection{Does multi-turn data generation help over single-turn data generation?} \label{sec:stepping-stone}

\newcommand{\chainhard}[1]{%
  \begin{tcolorbox}[colback=red!8, colframe=red!40, boxrule=0.3pt, arc=2pt,
    left=3pt, right=3pt, top=1pt, bottom=1pt, before skip=2pt, after skip=2pt]
  \textbf{\texttt{hard}} \quad \footnotesize\looseness=-1 #1
  \end{tcolorbox}}
\newcommand{\chainmedium}[1]{%
  \begin{tcolorbox}[colback=orange!8, colframe=orange!40, boxrule=0.3pt, arc=2pt,
    left=3pt, right=3pt, top=1pt, bottom=1pt, before skip=2pt, after skip=2pt]
  \textbf{\texttt{medium}} \quad \footnotesize\looseness=-1 #1
  \end{tcolorbox}}
\newcommand{\chaineasy}[1]{%
  \begin{tcolorbox}[colback=green!8, colframe=green!40, boxrule=0.3pt, arc=2pt,
    left=3pt, right=3pt, top=1pt, bottom=1pt, before skip=2pt, after skip=2pt]
  \textbf{\texttt{easy}} \quad \footnotesize\looseness=-1 #1
  \end{tcolorbox}}

\newcommand{\chaingroup}[4]{%
  \begin{tcolorbox}[colback=gray!4, colframe=gray!25, boxrule=0.4pt, arc=3pt,
    left=3pt, right=3pt, top=2pt, bottom=2pt, title={\small\textbf{#1}},
    before skip=4pt, after skip=4pt, boxsep=1pt]
  #2
  #3
  #4
  \end{tcolorbox}}

\begin{figure}[!ht]
\centering

\chaingroup{Chain Example 1}{%
\chainhard{Your task is to write an algorithm that, given a set of geographic nodes and directed edges between them, determines for each node the smallest angular sector that contains all bearings of its outgoing edges. The challenge lies in handling circular wrap-around (e.g., bearings near 0\textdegree{} and 359\textdegree) and producing exact degree spans efficiently.}}{%
\chainmedium{Your challenge is to write a function that, given a list of compass bearings (in degrees), finds the smallest angular sector that covers them all. The sector may need to wrap around the 0\textdegree/360\textdegree{} boundary, so careful handling of circular wrap-around is required. Return the sector size rounded to one decimal place.}}{%
\chaineasy{Your task is to write a function that, given two compass bearings (in degrees), returns the smallest angular distance between them. The distance must be between 0\textdegree{} and 180\textdegree, and you should round the answer to one decimal place. Think carefully about the circular nature of a compass when the two bearings lie on opposite sides of the 0\textdegree/360\textdegree{} line.}}

\chaingroup{Chain Example 2}{%
\chainhard{Your task is to implement \texttt{f} that computes the minimum number of node insertions and deletions needed to transform one ordered tree into another. Both trees are encoded as strings consisting solely of `(' and `)', where each pair represents a node. The challenge lies in handling arbitrary tree shapes, deriving the correct post-order indices, left-most leaf descendants, and keyroots, and then applying the Zhang-Shasha dynamic programming recurrence. Think carefully about the 1-based indexing used by the algorithm and ensure your implementation runs efficiently for the provided test cases.}}{%
\chainmedium{Your challenge is to implement \texttt{f} that, given two strings consisting only of `(' and `)', builds the corresponding ordered trees and returns the size (number of nodes) of the biggest \emph{rooted} sub-tree that appears identically at the top of both trees. In other words, start from the two roots and walk down the left-most children as long as the sub-trees match perfectly; sum the nodes you encounter. Think about how to parse the parenthesis representation into a tree and how to compare two sub-trees for exact structural equality. No randomisation or I/O is involved---just careful recursion and tree traversal.}}{%
\chaineasy{Your challenge is to write \texttt{f} that takes a single string made only of `(' and `)', builds the ordered tree encoded by that parenthesis representation, and returns the total number of nodes in the tree. Think about how each `(' creates a new node and each `)' finishes the current node, then recursively count the nodes. No I/O, randomness, or date/time operations are needed---just careful parsing and a simple recursion.}}

\caption{\textbf{Examples of \texttt{easy}--\texttt{medium}--\texttt{hard} question chains.}
Each chain illustrates a progression in task complexity within the same conceptual domain, derived by starting from a hard question and generating simpler variants that capture its core subproblems.
This reverse construction ensures that \texttt{easy} and \texttt{medium} questions correspond to meaningful components of the original hard task rather than trivial simplifications.}
\label{fig:example-chains}
\end{figure}

\begin{figure}[!h]
    \centering
    \hspace{-.7cm}
    \begin{subfigure}[t]{0.5\linewidth}
        \centering
        \includegraphics[width=\textwidth]{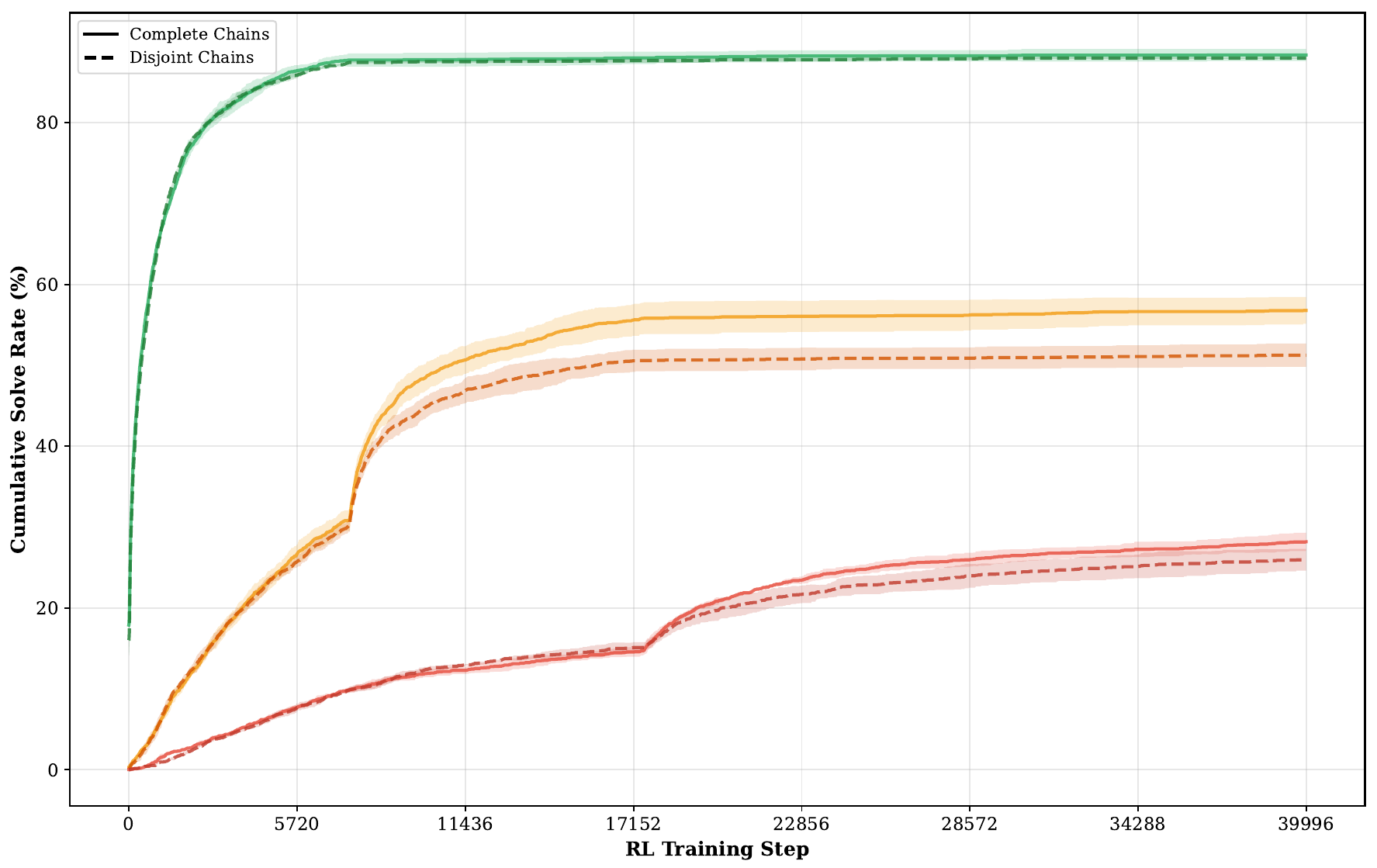}
        \caption{Hard curriculum}
        \label{fig:qwen-chains-hard-base}
    \end{subfigure}
    \hspace{.3cm}
    \begin{subfigure}[t]{0.5\linewidth}
        \centering
        \includegraphics[width=\textwidth]{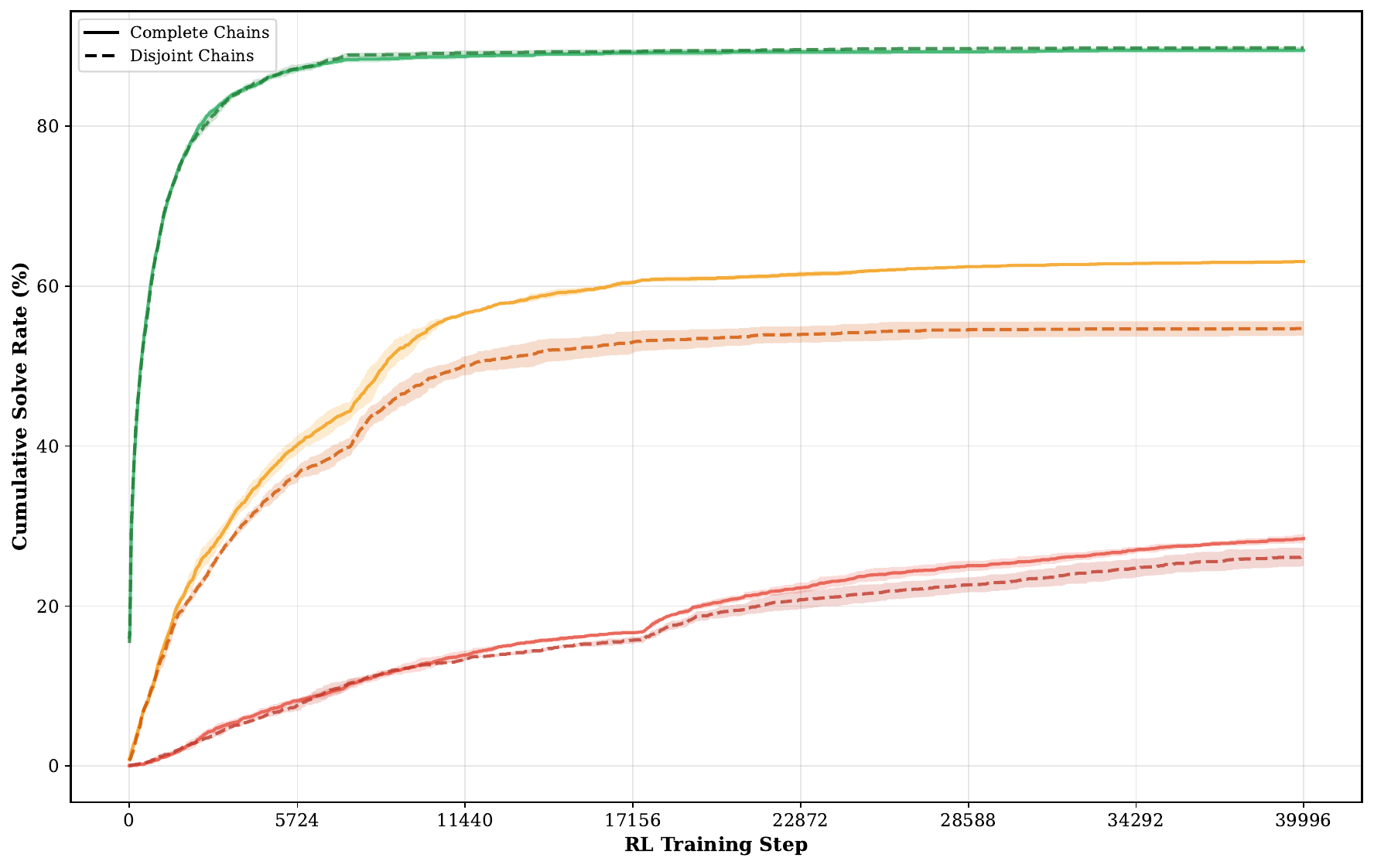}
        \caption{Soft curriculum}
        \label{fig:qwen-chains-hard-instruct}
    \end{subfigure}

\caption{\textbf{Qwen3-8B Base: Solve rates across \texttt{easy–medium–hard} splits under different curriculum strategies.}
(a)~Hard curriculum with abrupt transitions between difficulty levels. 
(b)~Soft curriculum with gradual transitions between difficulty levels. 
Curves depict cumulative solve rates for \textcolor{green!50!black}{\texttt{easy}}, \textcolor{orange!80!black}{\texttt{medium}}, and \textcolor{red!70!black}{\texttt{hard}} questions during RL training. Solid lines correspond to training on complete \texttt{easy–medium–hard} chains, while dashed lines show the disjoint-chain ablation.}

    \label{fig:qwen-chains-solve-rate}
\end{figure}

\begin{figure}[!h]
    \centering
    \begin{subfigure}[t]{1.0\textwidth}
        \centering
        \includegraphics[width=\textwidth]{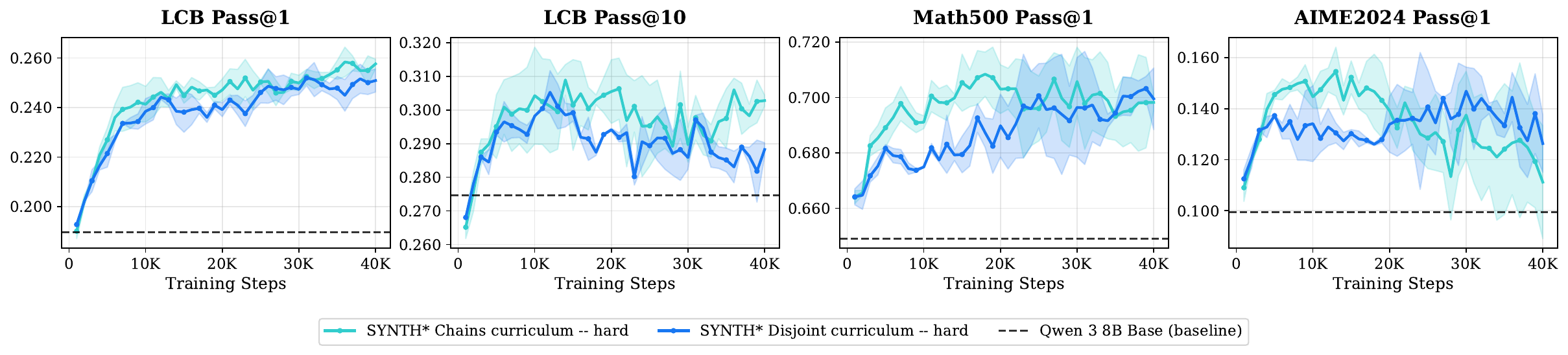}
        \caption{Hard curriculum: hard transitions between \texttt{easy}–\texttt{medium}–\texttt{hard} splits.}
        \label{fig:qwen-chains-hard-curr}
    \end{subfigure}

    \begin{subfigure}[t]{1.0\textwidth}
        \centering
        \includegraphics[width=\textwidth]{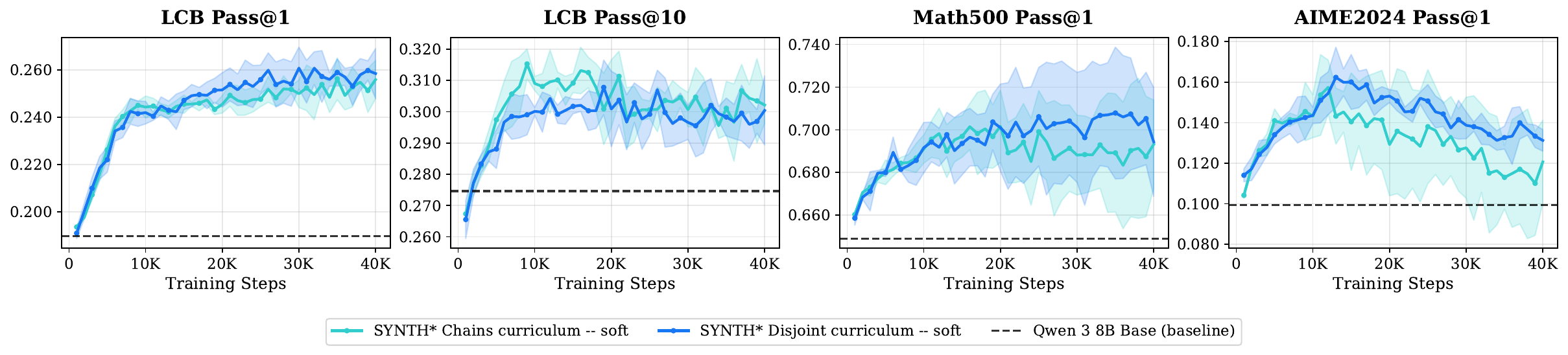}
        \caption{Soft curriculum: soft transitions between \texttt{easy}–\texttt{medium}–\texttt{hard} splits.}
        \label{fig:qwen-chains-soft-curr}
    \end{subfigure}

    \begin{subfigure}[t]{1.0\textwidth}
        \centering
        \includegraphics[width=\textwidth]{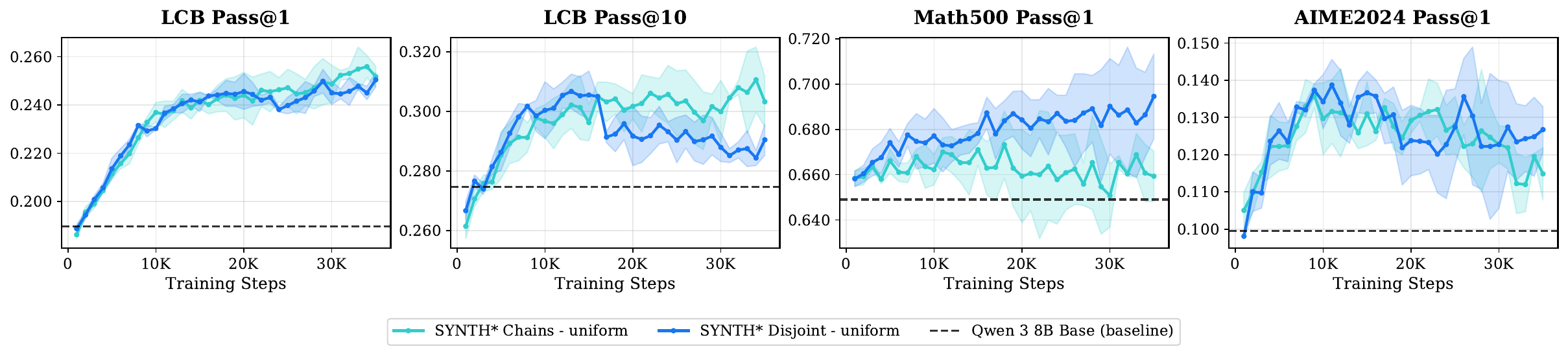}
        \caption{No curriculum: uniform sampling from \texttt{easy}–\texttt{medium}–\texttt{hard} splits throughout RL training.}
        \label{fig:qwen-chains-uniform}
    \end{subfigure}
\caption{\textbf{Qwen3-8B Base: RL on chains of \texttt{easy–medium–hard} questions with different curriculum strategies.}
Each row shows a different curriculum setup: 
(a)~hard transitions between difficulty levels, 
(b)~soft transitions, and 
(c)~no curriculum (uniform sampling). 
Results are shown for \texttt{LCB} (in-domain) and \texttt{Math500}, \texttt{AIME2024} (out-of-domain) with pass@1 and pass@10 metrics plotted over RL training steps.}
    \label{fig:qwen-chains-curricula}
\end{figure}

Multi-turn data generation improves both the quality and retention of synthetic problems. By allowing the teacher model to iteratively refine previous outputs, we reduce invalid generations caused by formatting errors or overly difficult tasks, 
increasing the fraction of viable problems by approximately fourfold after filtering compared to the same number of independent single-turn samples per seed.
Beyond this practical improvement, multi-turn generation introduces a second, conceptual advantage: it naturally produces stepping stones, \ie progressively refined task variants that can act as intermediate challenges during RL training.

Hard problems pose a significant exploration challenge under binary rewards, where positive feedback is only obtained for fully correct solutions. We therefore test whether the stepping-stone structure induced by multi-turn generation helps mitigate this challenge and facilitates more effective learning for RL.

\paragraph{\textbf{Building complete chains of \texttt{easy}-\texttt{medium}-\texttt{hard} questions.}}
To evaluate whether stepping stones help mitigate the hard-exploration challenge, we compare RL training on problem sets with and without explicit stepping-stone structure.
This process builds on the multi-turn generation pipeline (\sec{sec:multi-turn}), but reverses the direction: rather than generating problems across turns and filtering post-hoc, we start from \texttt{hard} questions and explicitly prompt the teacher to produce progressively easier variants, filtering between turns to ensure valid difficulty transitions (see \sec{sec:stepping-stone-details} for details on the filtering). This yields 1012 coherent \texttt{easy–medium–hard} question chains (3036 problems total) built around the same underlying task. Some examples are shown in \fig{fig:example-chains}.
Starting from hard questions ensures that we do not obtain artificially inflated difficulties (\eg by trivially modifying easy questions), and that the generated variants reflect meaningful variations in problem complexity.

In order to isolate the effect of explicit stepping-stone structure, we construct a baseline consisting of disjoint \texttt{easy–medium–hard} problems that share the same difficulty distribution as the chained data but are not derived from one another. Note that this is a strong baseline, as disjoint problems are drawn from independent seeds and thus offer greater overall diversity. In this baseline, the \texttt{easy}, \texttt{medium}, and \texttt{hard} splits are matched by their empirical student pass-rate histograms, ensuring comparable difficulty levels across datasets. While these problems do not form explicit chains, some implicit skill overlap may still occur, as certain questions may target similar problem types.

To ensure that the \texttt{easy} and \texttt{medium} counterparts are sufficiently reinforced before attempting harder variants, we replace uniform sampling across difficulty levels with curriculum-based scheduling.
We design two variants:
(1) Hard curriculum, where transitions between difficulty levels are sharp, \ie training begins almost exclusively on \texttt{easy} tasks, shifts predominantly to \texttt{medium}, and finally focuses on \texttt{hard} tasks.
(2) Soft curriculum, where \texttt{easy} and \texttt{medium} problems continue to appear throughout training, allowing for smoother transitions and mixed exposure.
The exact sampling weights for both curricula are provided in Appendix~\ref{app:curricula}.

As shown in \fig{fig:qwen-chains-solve-rate}, training on full chains with explicit stepping stones improves the model’s ability to solve \texttt{medium} and \texttt{hard} questions for both curriculum strategies.
However, when inspecting downstream benchmark performance, we observe nuanced differences across curricula (\fig{fig:qwen-chains-curricula}).

Although overall differences remain marginal, training with explicit stepping stones yields modest improvements under the hard curriculum, while effects under the soft and uniform setups are less consistent, particularly on out-of-domain benchmarks.

\citet{zhang2025grpo} show that for GRPO, if rewards are applied uniformly across problems regardless of their difficulty, models end up excessively optimizing simpler tasks while neglecting more challenging problems that require deeper reasoning. Similarly, \citet{qu2026pope} show that optimization on easier, already-solvable problems can actively inhibit progress on harder ones.
Echoing their findings, we also find that performance is often dominated by the \texttt{easy} split in our experiments, and mixing the splits throughout training hurts performance gains. 

In the context of stepping stones, however, this trade-off becomes a double-edged sword. Ideally, the model should continue to see the \texttt{easy} and \texttt{medium} variants of hard problems often enough to learn the underlying structure before attempting the hardest instances.
This effect is evident in \fig{fig:qwen-chains-solve-rate}, where the soft curriculum, which maintains exposure to the \texttt{easy} split later into training, achieves higher solve rates. Yet, mixing gradients from problems of varying difficulty can reduce the effectiveness of gradient updates across difficulty levels, resulting in solve-rate improvements that do not translate into downstream gains.
Overall, these results suggest that while explicit stepping stones offer mild benefits, their effectiveness depends strongly on curriculum design and the balance of rewards across difficulty levels.

\begin{takeawaybox}
Multi-turn generation introduces intermediate problem variants that can marginally improve learning under structured curricula.
Their benefit is most evident with hard curricula, while mixed-difficulty training can limit these gains by creating interference across tasks.
\end{takeawaybox}

\subsection{How much does the difficulty of questions matter for RL?}
\begin{figure}[!h]
    \centering
    \includegraphics[width=1.0\textwidth]{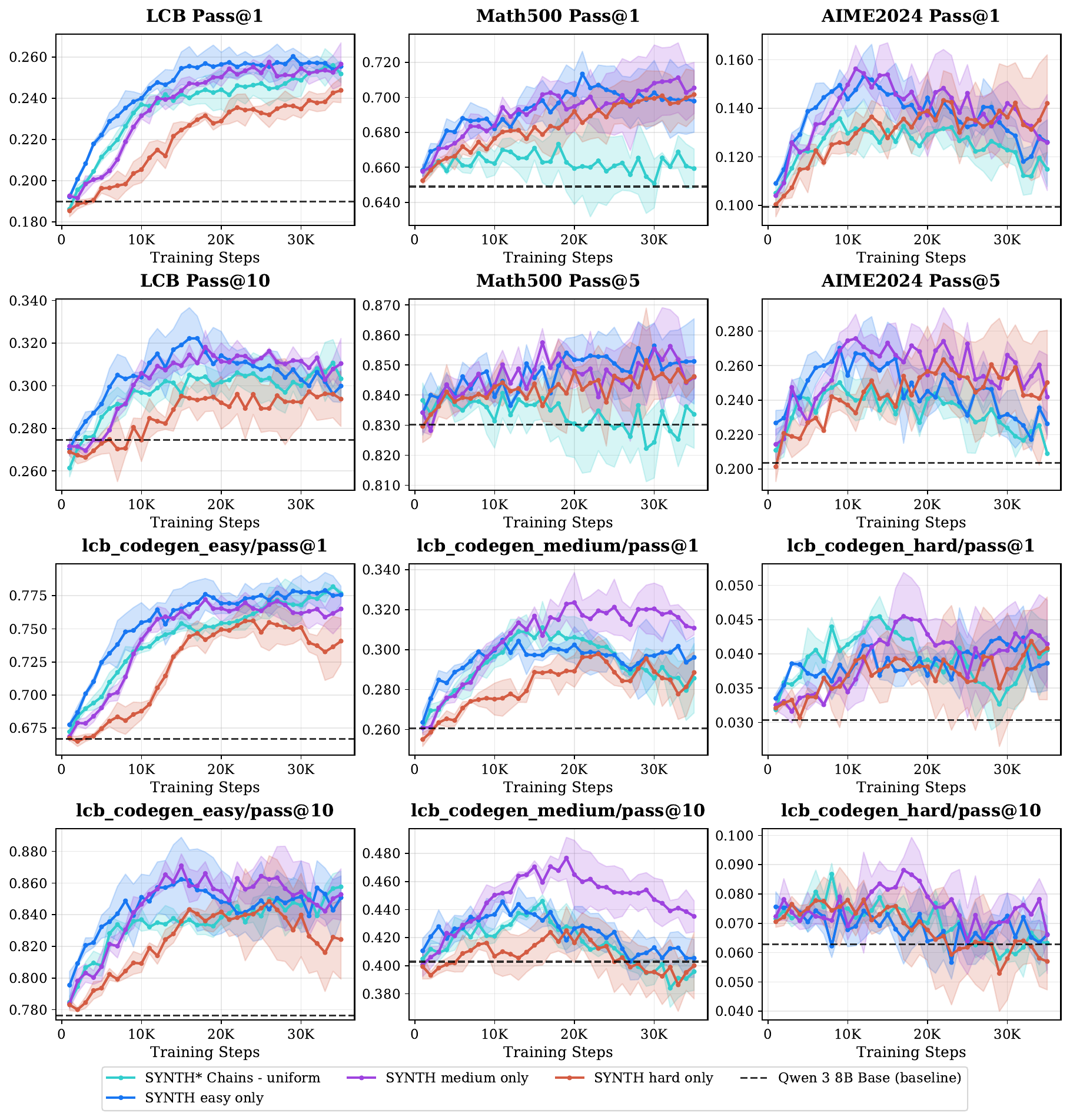}
    
    \caption{\textbf{Qwen3-8B Base: RL on \texttt{easy}, \texttt{medium}, \texttt{hard} splits.}
Comparison of RL training when restricted to questions of a single difficulty level (3036 per level) across \texttt{easy}, \texttt{medium}, and \texttt{hard} subsets, with the uniform chain baseline (3036 total, 1012 per level) for reference. Performance shown on in-domain (code: \texttt{LCB}) and out-of-domain (math: \texttt{Math500}, \texttt{AIME2024}) benchmarks across 3 seeds.}
    \label{fig:qwen-e-m-h}
\end{figure}

So far, we have operated under the assumption that solving harder questions provides a more valuable learning signal than solving easier ones.
We test this hypothesis by examining how question difficulty shapes RL dynamics.
Instead of using mixed chains of 3036 questions across difficulty levels (\texttt{easy}, \texttt{medium}, \texttt{hard}; 1012 each), we train models exclusively on questions from a single difficulty level (3036 per difficulty level) (\fig{fig:qwen-e-m-h}).
Training on \texttt{easy} questions yields early gains and faster convergence, as expected.
However, training on \texttt{medium} questions achieves comparable or better overall performance despite slightly slower convergence.
Notably, models trained on \texttt{medium}-level questions perform better on the \texttt{LCBv5-medium} and \texttt{LCBv5-hard} splits.
This suggests that the apparent early gains from \texttt{easy} questions may reflect overfitting to simpler problem types.
Training on the \texttt{hard} split, in contrast, results in much slower convergence due to the sparsity of positive rewards.
Although models trained on \texttt{hard} questions remain competitive on out-of-domain math benchmarks, their performance lags behind on \texttt{LCB}.
These results highlight the importance of training on non-trivial problems that still provide a dense enough learning signal for effective RL optimization.

\begin{takeawaybox}
Training on \texttt{medium}-level questions offers the best balance between convergence speed and generalization,
while \texttt{easy} questions risk overfitting and \texttt{hard} ones suffer from sparse rewards.
\end{takeawaybox}

\subsection{Does the curriculum matter?}
\begin{figure}[!hb]
    \centering
    \includegraphics[width=1.0\textwidth]{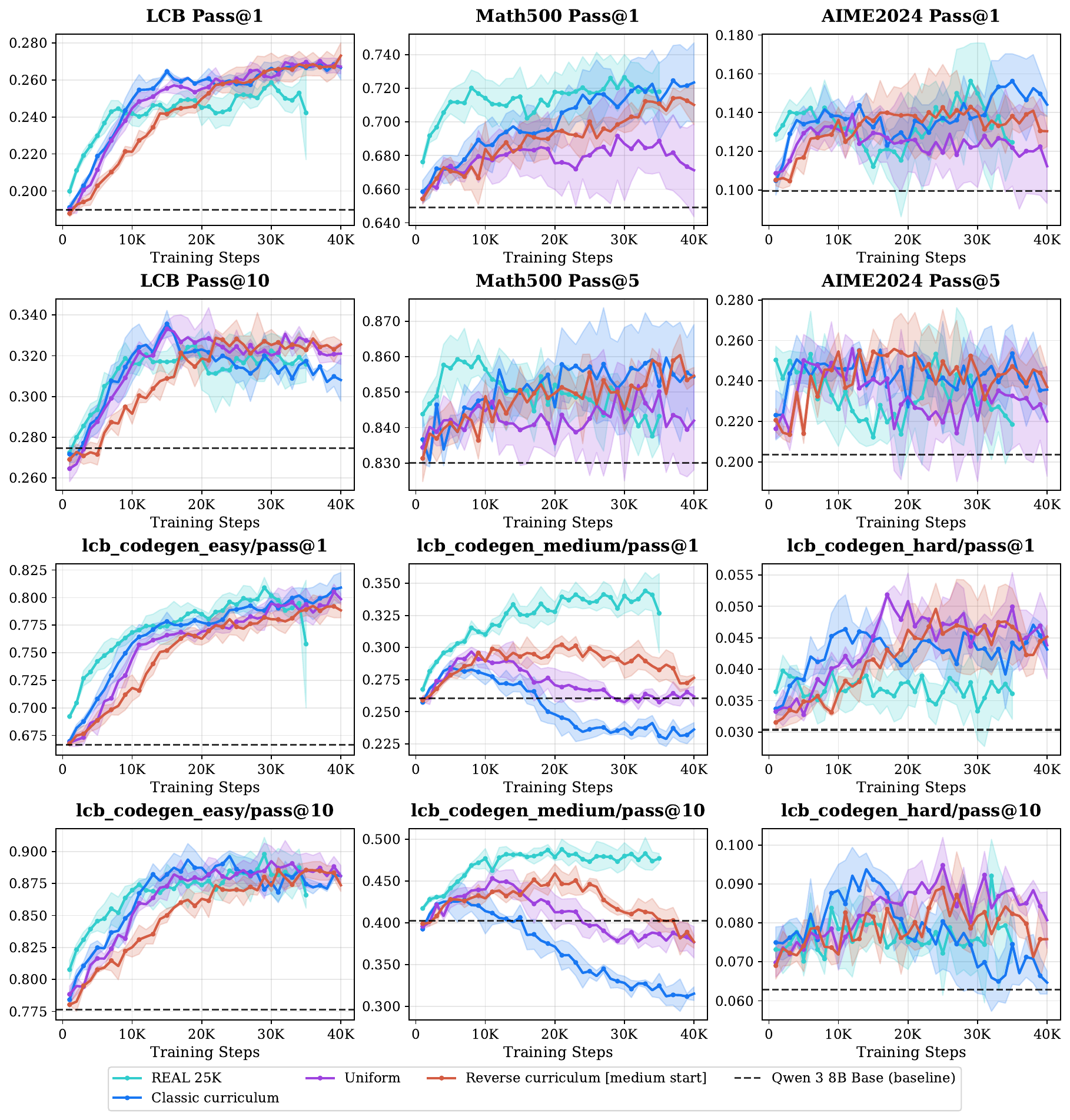}
\caption{\textbf{Comparing different curriculum strategies in Qwen3-8B Base.}
The figure compares RL training with different curriculum strategies over extended difficulty splits (\texttt{easy}, \texttt{easy–medium}, \texttt{medium}, \texttt{hard}). 
Curves show pass@1 and pass@10 on the in-domain benchmark (\texttt{LCB}, including \texttt{LCBv5}-\texttt{easy}, \texttt{medium}, and \texttt{hard} splits) and pass@1 and pass@5 on the out-of-domain benchmarks (\texttt{Math500}, \texttt{AIME2024}), averaged across three seeds. 
Reverse curricula starting from \texttt{medium}-level questions mitigate overfitting to \texttt{easy} tasks and maintain stronger performance on harder splits.}

    \label{fig:qwen-reverse-curr-medium}
\end{figure}
 
\begin{figure}[!hb]
    \centering
    \includegraphics[width=1.0\textwidth]{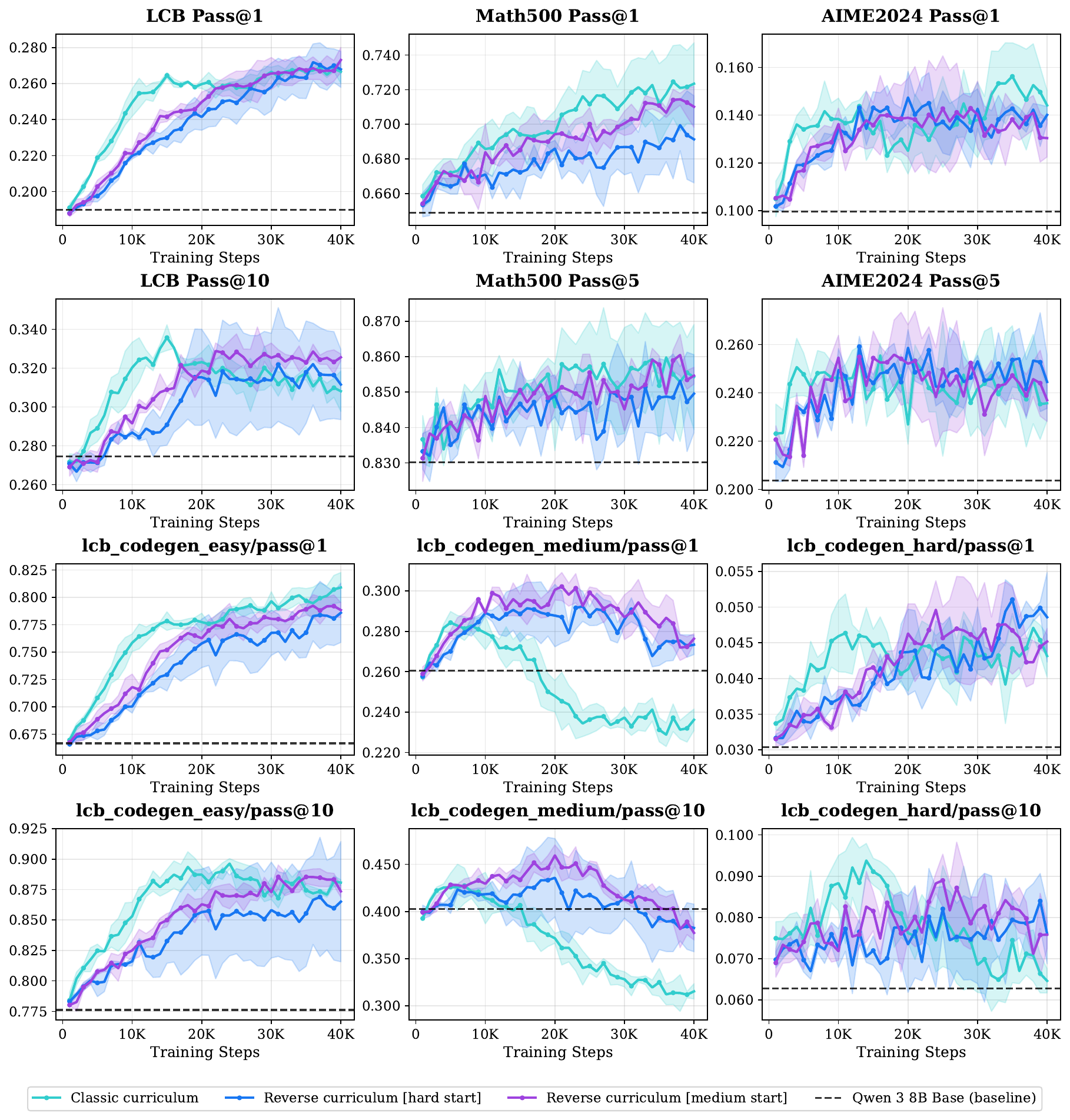}
\caption{\textbf{Reversed curriculum experiments in Qwen3-8B Base.}
The figure compares RL training with different reverse curriculum strategies over extended difficulty splits (\texttt{easy}, \texttt{easy–medium}, \texttt{medium}, \texttt{hard}). 
We evaluate two variants: starting with the \texttt{medium} split (“medium-start”) and starting with the \texttt{hard} split (“hard-start”). 
Curves show pass@1 and pass@10 on the in-domain benchmark (\texttt{LCB}, including \texttt{LCBv5}-\texttt{easy}, \texttt{medium}, and \texttt{hard} splits) and pass@1 and pass@5 on the out-of-domain benchmarks (\texttt{Math500}, \texttt{AIME2024}), averaged across three seeds. 
The medium-start reverse curriculum shows faster convergence and lower variance across seeds compared to the hard-start variant.}

    \label{fig:qwen-reverse-curr-hard}
\end{figure}

As observed in our earlier experiments with stepping-stones, the choice of curriculum can influence RL dynamics, although the performance differences were not always significant. Here, we further investigate the extent to which the curriculum schedule affects training outcomes.

In these experiments, we expand the dataset by adopting a broader binning for \texttt{easy}, \texttt{medium}, and \texttt{hard} questions. To retain more data, we do not homogenize the number of examples per split.

We test the following hypothesis: conventional curricula typically progress from easier to harder problems ($\mathtt{easy} \rightarrow \mathtt{medium} \rightarrow \mathtt{hard}$). However, as the model’s entropy naturally decreases during RL training, starting with easy problems may waste exploration capacity on tasks that the model can already solve, potentially reducing its ability to solve harder questions later. Motivated by this intuition, we evaluate a \emph{reverse curriculum}, in which training proceeds from harder to easier problems ($\mathtt{hard} \rightarrow \mathtt{medium} \rightarrow \mathtt{easy}$).
Given that \texttt{medium}-level questions previously showed strong generalization performance (\fig{fig:qwen-e-m-h}), we also experiment with a \emph{medium-start reverse curriculum} ($\mathtt{medium} \rightarrow \mathtt{easy\text{-}medium}$). Exact sampling schedules for all curricula are provided in \tab{tab:classic_reverse_curricula}.
The difficulty bins and corresponding data sizes are summarized in \tab{tab:curriculum-bins} (see Appendix~\sec{app:curriculum-bins}).

Results in \fig{fig:qwen-reverse-curr-medium} show that: 
(1) uniform sampling across difficulty levels is suboptimal, as performance is largely driven by the \texttt{easy} split; 
(2) the classic curriculum yields faster convergence and stronger performance on easier problems, but tends to underperform on more challenging splits; and 
(3) the reverse curriculum achieves a more balanced outcome, improving stability and yielding higher scores on \texttt{LCBv5-medium}, though gains on \texttt{LCBv5-hard} remain limited. 
These findings suggest that reversing the curriculum order can help maintain exploration and prevent overfitting to trivial cases without sacrificing convergence speed.
Notably, in the \texttt{medium}-start reverse curriculum: once the training transitions towards the \texttt{easy–medium} split (after $\sim$20K steps), both \texttt{medium} and \texttt{hard} performance begin to plateau, and in some cases degrade slightly.
This suggests that while reverse curricula help maintain exploration early on, continued exposure to easier tasks later in training may again bias optimization toward simpler problems.

We also evaluate a reverse curriculum that starts directly from the \texttt{hard} split, which poses a challenging exploration problem early in training (\fig{fig:qwen-reverse-curr-hard}). 
This setup leads to slightly slower convergence compared to the medium-start variant, as the model initially struggles to obtain positive rewards on the hardest problems. 
More notably, we observe substantially higher variance across random seeds, reflecting the sensitivity of early learning to rare successful samples when rewards are sparse. 
These results suggest that while reverse curricula can encourage exploration, beginning from excessively difficult tasks may destabilize optimization.

\begin{takeawaybox}
Uniform and classic curricula tend to overfit to \texttt{easy}-level questions.
Reverse curricula that begin from \texttt{medium} tasks encourage broader exploration and more stable learning, though their advantage over training solely on \texttt{medium}-level data requires further investigation.
\end{takeawaybox}

\subsection{How does environment diversity affect RL performance?}

\begin{figure}[!h]
    \centering
    \includegraphics[width=1.0\textwidth]{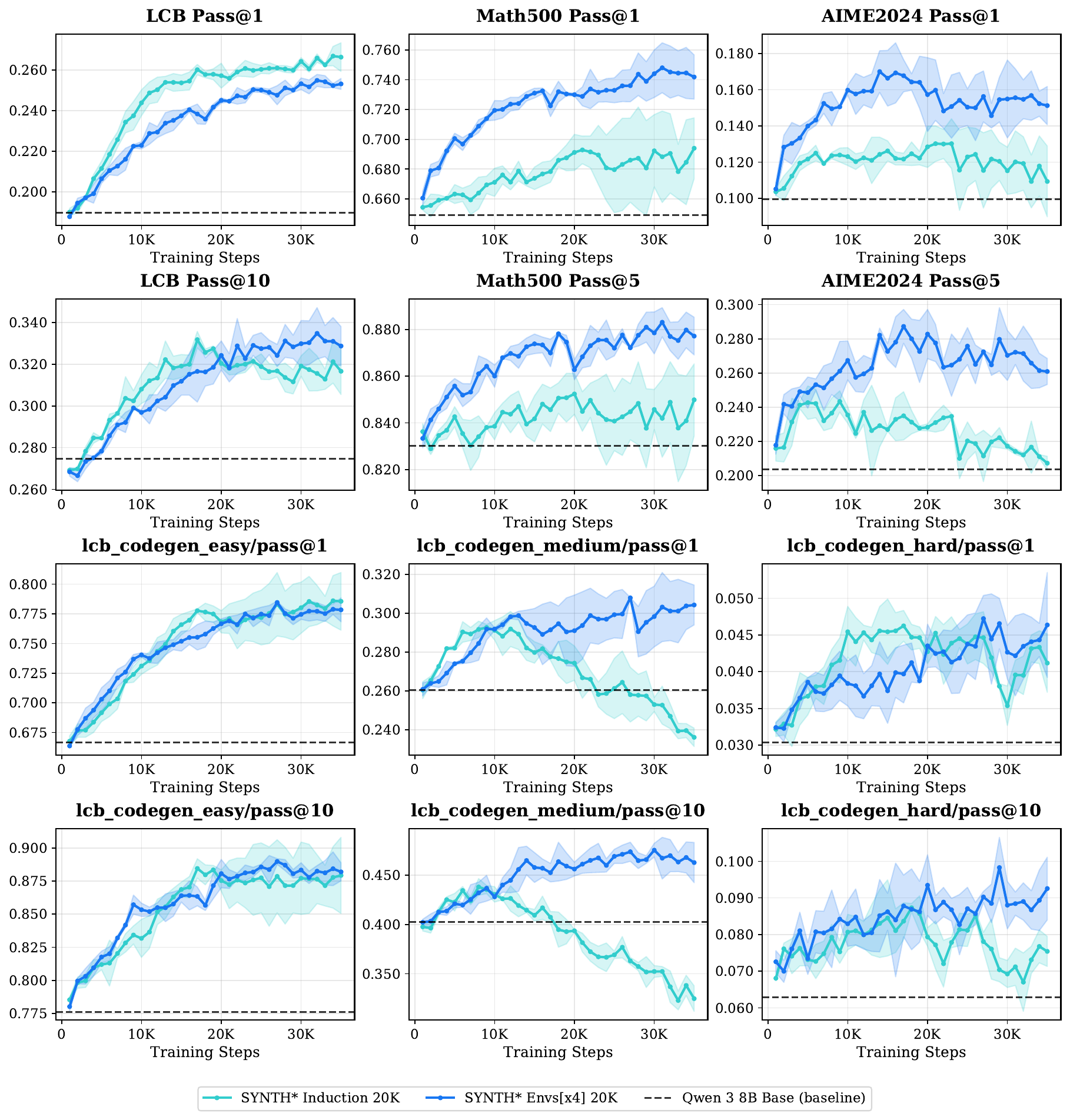}
\caption{\textbf{Scaling the number of RL environments in Qwen3-8B Base.}
The figure compares RL training on 20K synthetic problems from a single environment (\texttt{induction}) with training on an equal total of 20K synthetic problems distributed across four environments: \texttt{induction}, \texttt{abduction}, \texttt{deduction}, and \texttt{fuzzing}. 
Curves show pass@1 and pass@10 on in-domain (\texttt{LCB}) and pass@1 and pass@5 out-of-domain (\texttt{Math500}, \texttt{AIME2024}) benchmarks throughout RL training. 
Distributing the data budget across environments improves out-of-domain generalization and yields higher pass@10 scores in-domain, while reducing overfitting on the \texttt{easy} split.}

    \label{fig:qwen-env-scaling}
\end{figure}

\begin{figure}[!h]
    \centering
    \includegraphics[width=1.0\textwidth]{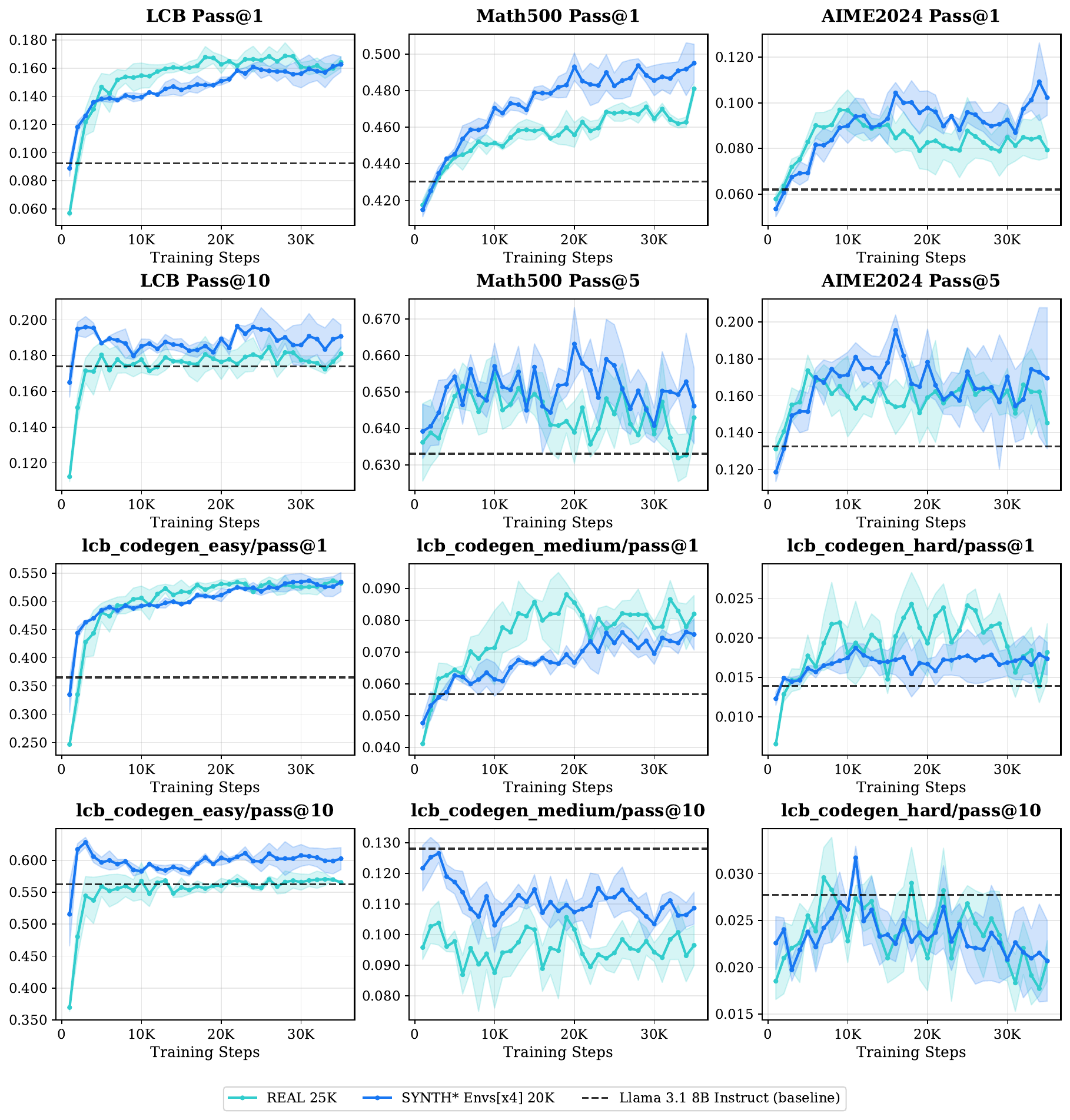}
\caption{\textbf{Scaling the number of RL environments in Llama3.1-8B Instruct compared to real data.}
The figure compares RL training on 25K real code-contest problems with training on a total of 20K synthetic problems distributed across four environments: \texttt{induction}, \texttt{abduction}, \texttt{deduction}, and \texttt{fuzzing}. 
Curves show pass@1 and pass@10 on the in-domain benchmark (\texttt{LCB}) and pass@1 and pass@5 on out-of-domain benchmarks (\texttt{Math500}, \texttt{AIME2024}) throughout RL training. 
Despite using less data, training across multiple synthetic environments improves out-of-domain generalization and yields higher in-domain pass@10 scores compared to real-only training.}
    \label{fig:llama-env-scaling}
\end{figure}

So far, we have seen that increasing data diversity improves performance. The gains observed from synthetic data augmentation can also be viewed as a form of diversity injection: although the \texttt{induction} environment shares the core principles of typical coding puzzles, it differs in input–output structure and question templates. 

We now ask: given a fixed RL problem budget, is it more beneficial to allocate it entirely to one environment with sufficient internal diversity, or to distribute it across multiple environments to introduce structural variation? To test this, we compare two settings: (1) RL training on 20K problems from the \texttt{induction} environment, and (2) training on 5K problems each from four distinct environments: \texttt{induction}, \texttt{abduction}, \texttt{deduction}, and \texttt{fuzzing}.

As shown in \fig{fig:qwen-env-scaling}, distributing the data budget across environments yields significant improvements on out-of-domain benchmarks. Even in in-domain evaluations, where pass@1 slightly lags behind the single-environment baseline, the multi-environment setup surpasses it in pass@10. Notably, unlike pure \texttt{induction} training, we observe no overfitting on the \texttt{easy} split, which otherwise harms performance on \texttt{medium}. The same trends hold when compared to RL training on 25K real code-contest questions (\fig{fig:qwen-env-scaling-wreal}).
\fig{fig:llama-env-scaling} shows a similar trend for Llama3.1-8B Instruct: training on 20K problems distributed across the four environments outperforms RL on 25K real code-contest questions, with higher out-of-domain generalization and improved pass@10 scores on \texttt{LCB}. 

\begin{takeawaybox}
The number of RL environments can be viewed as an additional scaling axis:
\begin{itemize}
\item Our experiments show that the diversity injected via increasing the number of RL environments leads to performance gains in pass@k scores, with less overfitting on easy tasks in \texttt{LCBv5}.
\item Different environments can target/aid different skill sets, as shown in our improved performance on out-of-domain math benchmarks when training across multiple environments.
\end{itemize}
\end{takeawaybox}

\section{Discussion}

Our experiments highlight several factors that influence the effectiveness of RL fine-tuning in the coding domain.
First, synthetic augmentation of real coding problems consistently improves both convergence speed and final performance.
The gains stem not merely from additional data but from the diversity introduced by varying input–output structures and templates.
Synthetic data generated from random \texttt{starcoderdata} snippets perform comparably to those seeded with real questions, underscoring the importance of diversity.
When used alone, synthetic RL problems remain competitive, though their difficulty distribution requires explicit control to avoid overfitting on easy problems.

Second, multi-turn data generation provides an additional advantage by increasing the proportion of valid problems generated and implicitly introducing intermediate problem variants that can serve as stepping stones during RL.
While these stepping stones offer modest benefits under structured curricula, their impact depends strongly on how difficulty levels are mixed during training.
Hard curricula benefit most, whereas excessive interleaving of easy and hard problems introduces interference that weakens the benefits of stepping stones.

Third, our results show that problem difficulty and curriculum design jointly determine the efficiency of RL optimization: training on \texttt{medium}-level problems offers the best balance between convergence speed and generalization, while \texttt{easy} problems lead to overfitting and \texttt{hard} ones suffer from sparse rewards.
Results indicate that the apparent early gains from \texttt{easy} problems in RL data mixes provide limited long-term benefit and may even hinder generalization. Reverse curricula that begin from \texttt{medium} tasks mitigate early overfitting, though their advantage over training solely on \texttt{medium}-level data requires further investigation.
 
Finally, we find that scaling the number of RL environments serves as a complementary axis of improvement.
Distributing the data budget across diverse environments (\texttt{induction}, \texttt{abduction}, \texttt{deduction}, \texttt{fuzzing}) improves out-of-domain generalization and reduces overfitting on easy in-domain splits.
These results suggest that environment diversity can play a role comparable to data scale in shaping robust RL behavior.

\arxiv{\bibliographystyle{assets/plainnat}}
\bibliography{refs} 

\iclr{
\bibliographystyle{iclr2026_conference}
}

\newpage

\appendix  

\section{Extended Results for RL with Synthetic Data}

\begin{figure}[!h]
    \centering
    \includegraphics[width=1.0\textwidth]{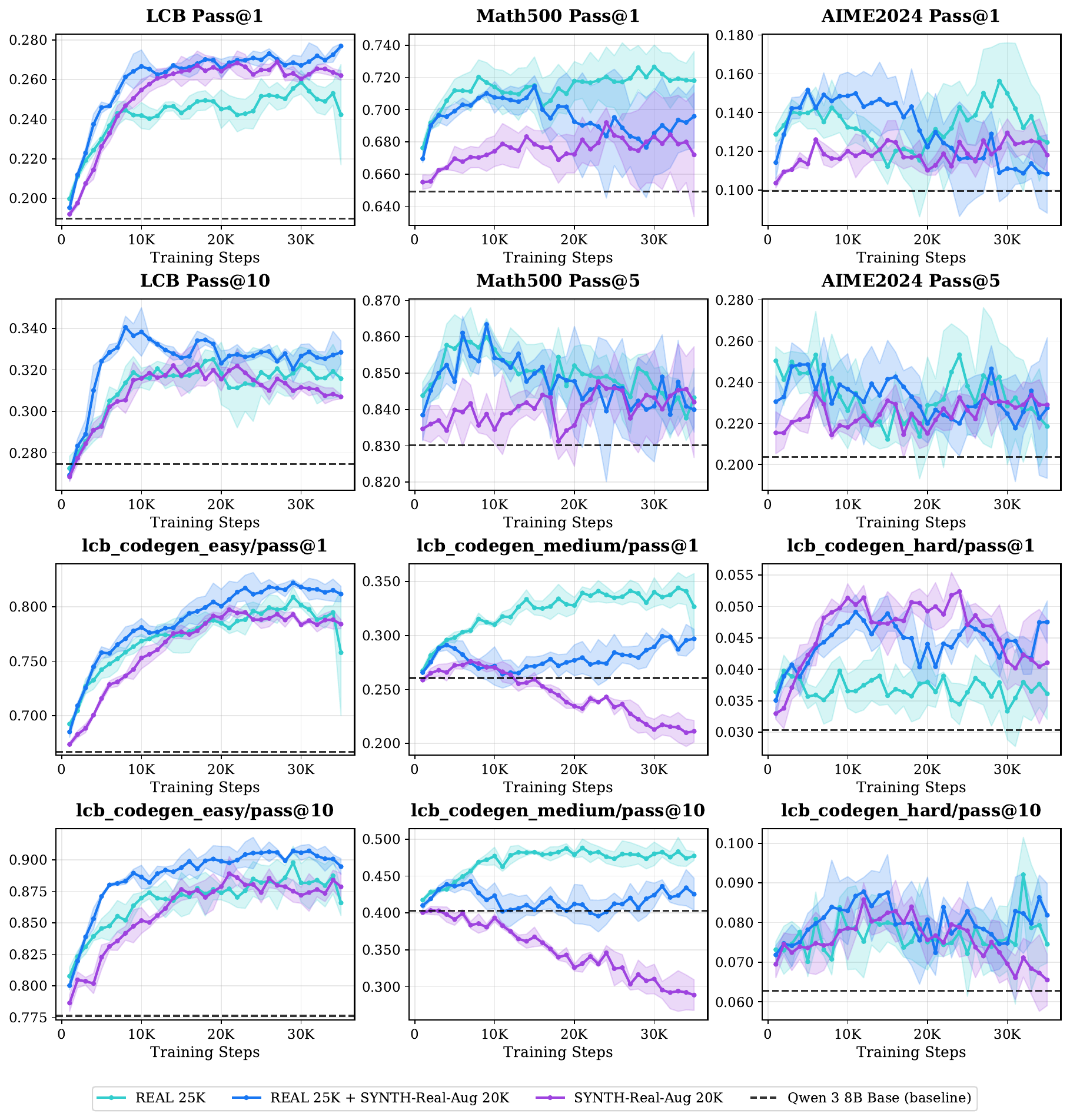}
\caption{\textbf{RL with synthetic data augmentation seeded from real problems in Qwen3-8B Base.}
RL training on 25K real code-contest questions (baseline) versus the same data augmented with 20K synthetic \texttt{induction} problems generated using real questions as inspiration seeds (\texttt{Real 25K + SYNTH-Real-Aug 20K}), as well as training solely on the synthetic partition (\texttt{SYNTH-Real-Aug 20K}). 
Performance shown on in-domain (\texttt{LCB}) and out-of-domain (\texttt{Math500}, \texttt{AIME2024}) benchmarks.}
    \label{fig:qwen-synth-aug-real-full}
\end{figure}

\begin{figure}[!h]
    \centering
    \includegraphics[width=1.0\textwidth]{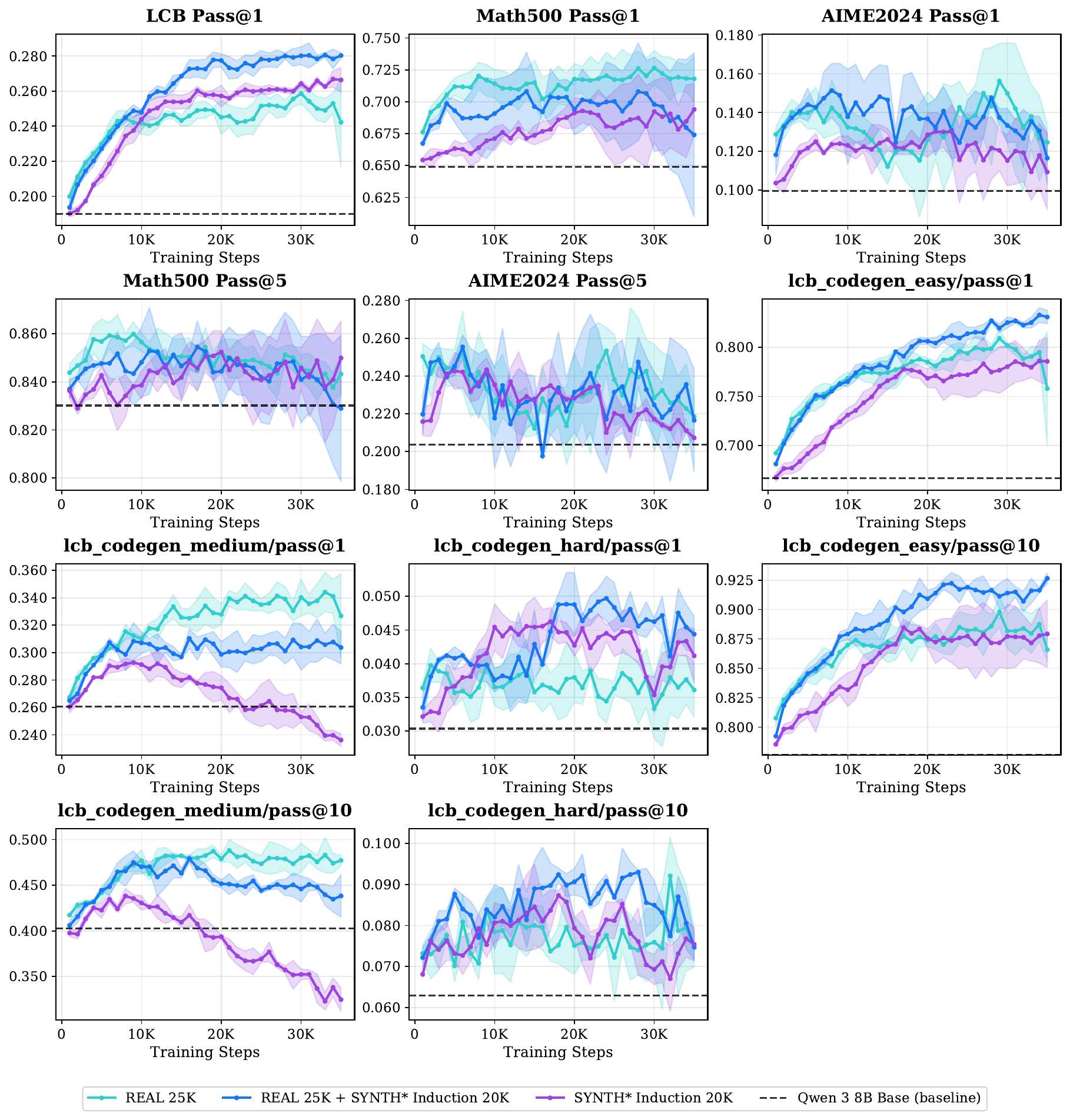}
\caption{\textbf{RL with synthetic data augmentation seeded from random code snippets in Qwen3-8B Base.}
RL training on 25K real code-contest questions (baseline) versus the same data augmented with 20K synthetic \texttt{induction} problems generated using random snippets from \texttt{starcoderdata} (\texttt{Real 25K + SYNTH* Induction 20K}), as well as training solely on the synthetic partition (\texttt{SYNTH* Induction 20K}). 
Performance shown on in-domain (\texttt{LCB}) and out-of-domain (\texttt{Math500}, \texttt{AIME2024}) benchmarks.}

    \label{fig:qwen-synth-aug-starcoder-full}
\end{figure}

\begin{figure}[!h]
    \centering
    \includegraphics[width=1.0\textwidth]{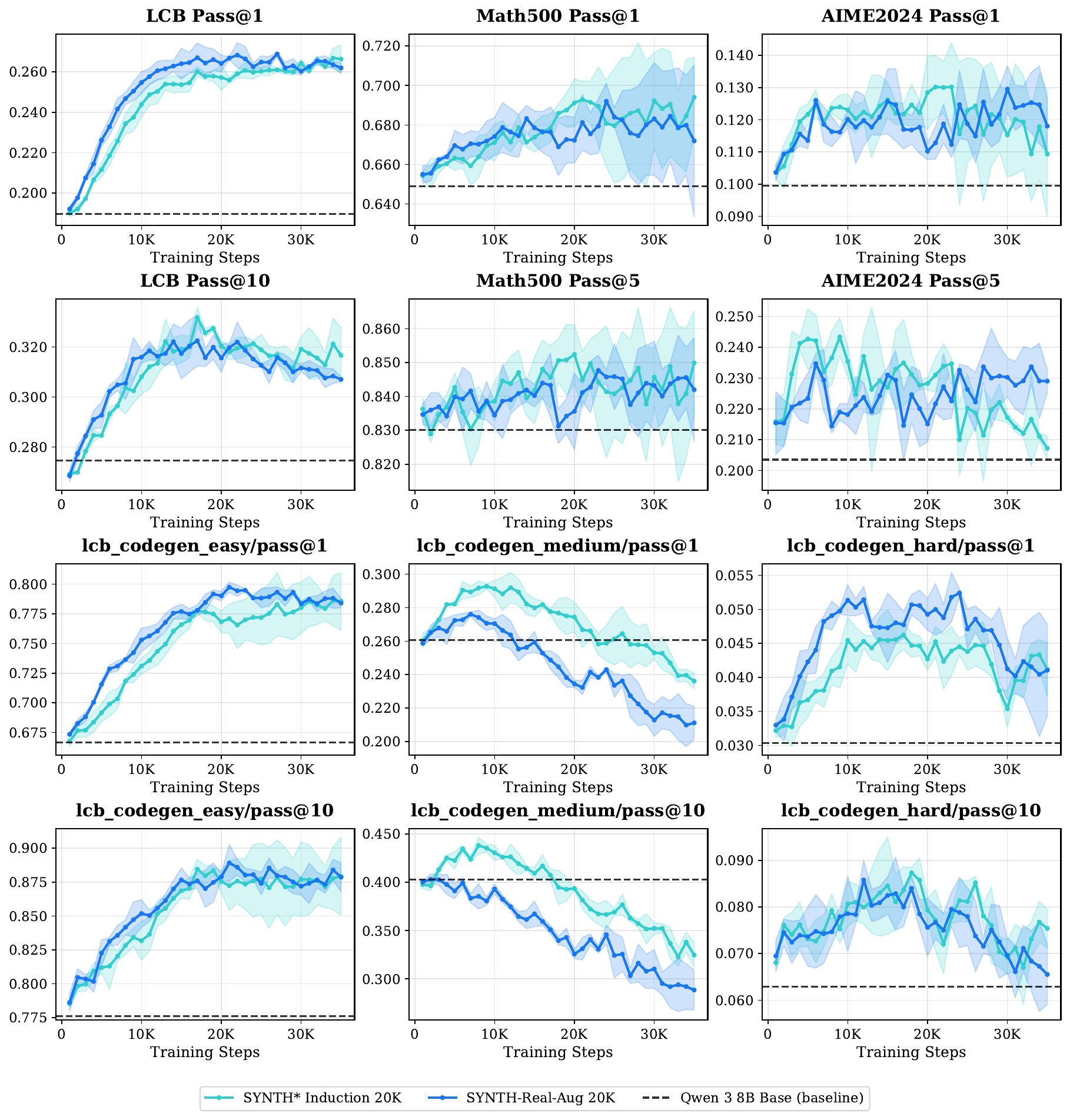}
\caption{\textbf{RL with pure synthetic data in Qwen3-8B Base.}
Instead of augmenting the 25K real code-contest questions, RL is performed solely on synthetic \texttt{induction} problems generated with two seeding strategies: 
(1) real questions as inspiration seeds (\texttt{SYNTH-Real-Aug}), and 
(2) random code snippets from \texttt{starcoderdata} (\texttt{SYNTH*-Aug}, 3 seeds). 
}
    \label{fig:qwen-synth-aug-full}
\end{figure}

\begin{figure}[!h]
    \centering
    \includegraphics[width=1.0\textwidth]{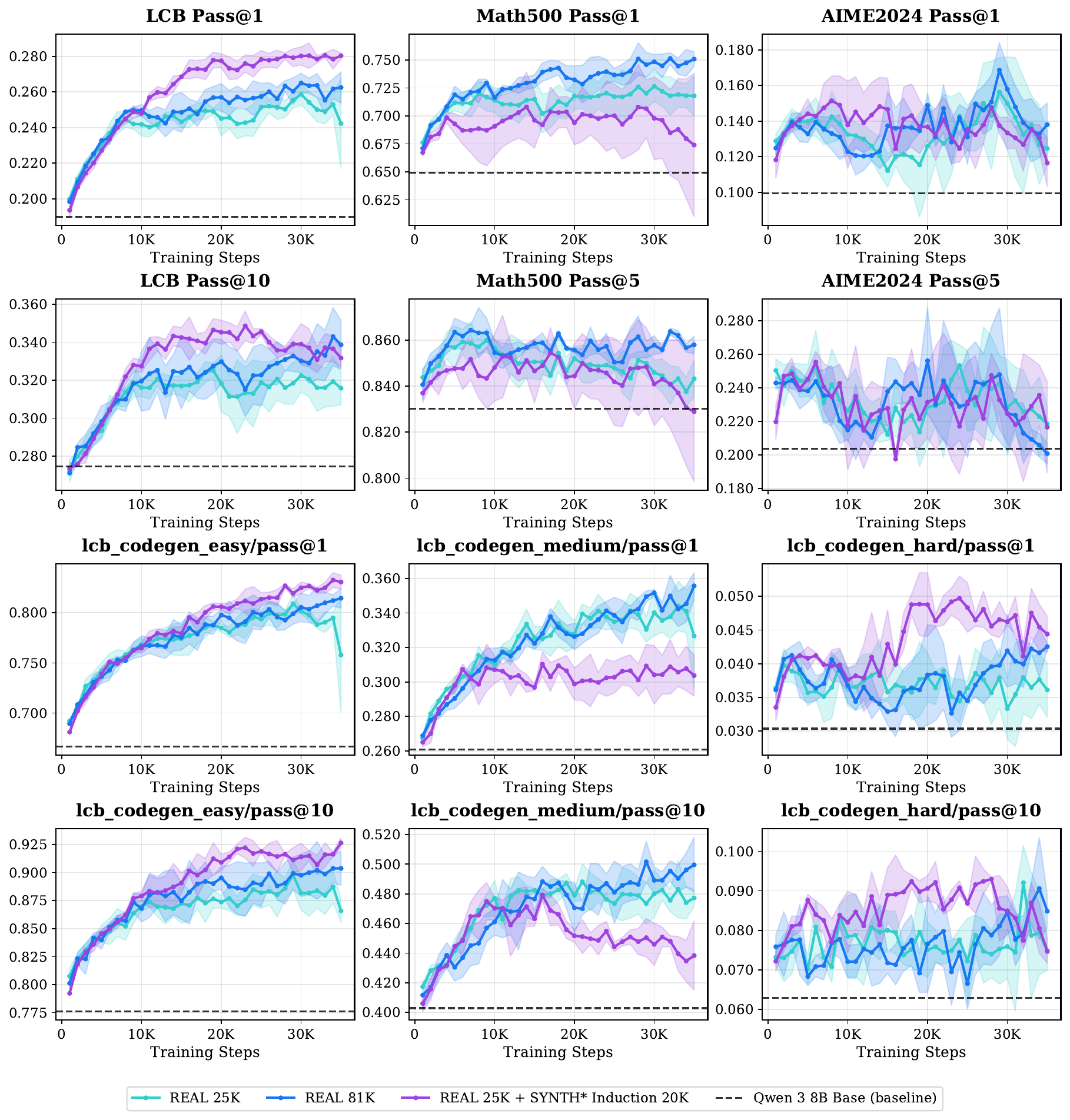}
\caption{\textbf{Scaling with real data vs. synthetic augmentation in Qwen3-8B Base.}
Comparison of RL training on 25K and 81K real code-contest problems, and on 25K real data augmented with 20K synthetic problems generated from random \texttt{starcoderdata} snippets (\texttt{Real 25K + SYNTH*-Aug 20K}). 
Performance is shown on in-domain (\texttt{LCB}) and out-of-domain (\texttt{Math500} and \texttt{AIME2024}) benchmarks throughout training. 
Synthetic augmentation yields additional gains over the 81K real-data baseline across all \texttt{LCB} splits except \texttt{medium}.}
    \label{fig:qwen-synth-aug-and-scale-full}
\end{figure}

Here, we provide extended results for the synthetic data experiments discussed in Section~\ref{sec:results_synthetic}. 
\Fig{fig:qwen-synth-aug-real-full}–\Fig{fig:qwen-synth-aug-starcoder-full} show detailed in-domain and out-of-domain performance for augmentations seeded with real questions and random code snippets, respectively, while \Fig{fig:qwen-synth-aug-full} compares RL training using purely synthetic datasets. 
\Fig{fig:qwen-synth-aug-and-scale-full} further contrasts scaling real data (25K vs.~81K problems) with augmenting real data using 20K synthetic problems, showing that synthetic augmentation can outperform even the larger real-data baseline.

\FloatBarrier

\FloatBarrier

\section{Stepping Stones with Easy-Medium-Hard Chains}\label{sec:stepping-stone-details}
To construct the \texttt{easy}, \texttt{medium}, and \texttt{hard} categories used for building question chains, we bin problems according to their empirical student pass rates, computed over 32 solution attempts per problem.
Problems with average pass rates in the range $0.81$–$0.91$ are labeled as \texttt{easy}, those between $0.41$–$0.59$ as \texttt{medium}, and those between $0.05$–$0.16$ as \texttt{hard}.
These thresholds ensure a clear separation between difficulty levels.
\vspace{-.3cm}
\subsection{Curriculum Strategies for Easy–Medium–Hard Chains} \label{app:curricula}
We provide details of the hard and soft curriculum strategies used in \sec{sec:stepping-stone} and \fig{fig:qwen-chains-curricula}.
Sampling probabilities for \texttt{easy}, \texttt{medium}, and \texttt{hard} tasks in the soft and hard curricula are listed below.

\begin{table}[h!]
\centering
\caption{Sampling weights across training stages for soft and hard curricula.}
\vspace{-1em}
\label{tab:curricula}
\begin{adjustbox}{width=0.35\textwidth,center}
\begin{tabular}{lccc}
\toprule
\textbf{Stage (RL steps)} & \textbf{Easy} & \textbf{Medium} & \textbf{Hard} \\
\midrule
\multicolumn{4}{l}{\textbf{Soft curriculum}} \\
0–7,500      & 0.80 & 0.15 & 0.05 \\
7,500–17,500 & 0.15 & 0.80 & 0.05 \\
17,500–30,000 & 0.15 & 0.40 & 0.45 \\
30,000–40,000 & 0.05 & 0.15 & 0.80 \\
\midrule
\multicolumn{4}{l}{\textbf{Hard curriculum}} \\
0–7,500      & 0.90 & 0.05 & 0.05 \\
7,500–17,500 & 0.05 & 0.90 & 0.05 \\
17,500–40,000 & 0.05 & 0.05 & 0.90 \\
\bottomrule
\end{tabular}
\end{adjustbox}
\end{table}

\vspace{-.6cm}
\subsection{Extended Results}
\begin{figure}[!h]
    \centering
    \includegraphics[width=.75\textwidth]{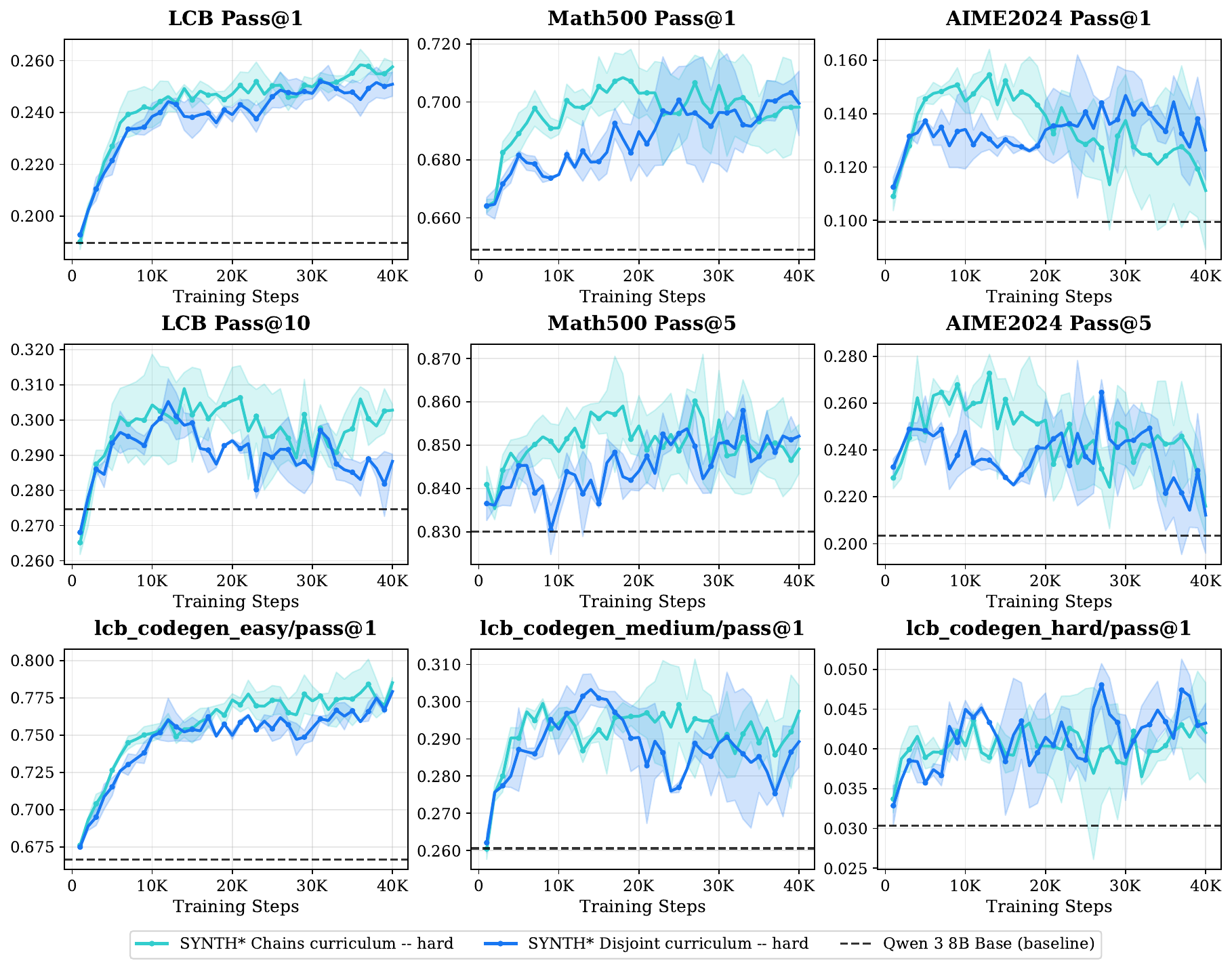}
\caption{\textbf{Chains with hard curriculum in Qwen3-8B Base.}
Extended results for RL training on \texttt{easy–medium–hard} question chains using the hard curriculum schedule. 
Shown are in-domain results (\texttt{LCB}, including easy, medium, and hard splits) and out-of-domain results (\texttt{Math500}, \texttt{AIME2024}) with pass@1, pass@5/pass@10 metrics plotted over RL training steps.}

    \label{fig:qwen-chains-hard}
\end{figure}

\begin{figure}[!h]
    \centering
    \includegraphics[width=.75\textwidth]{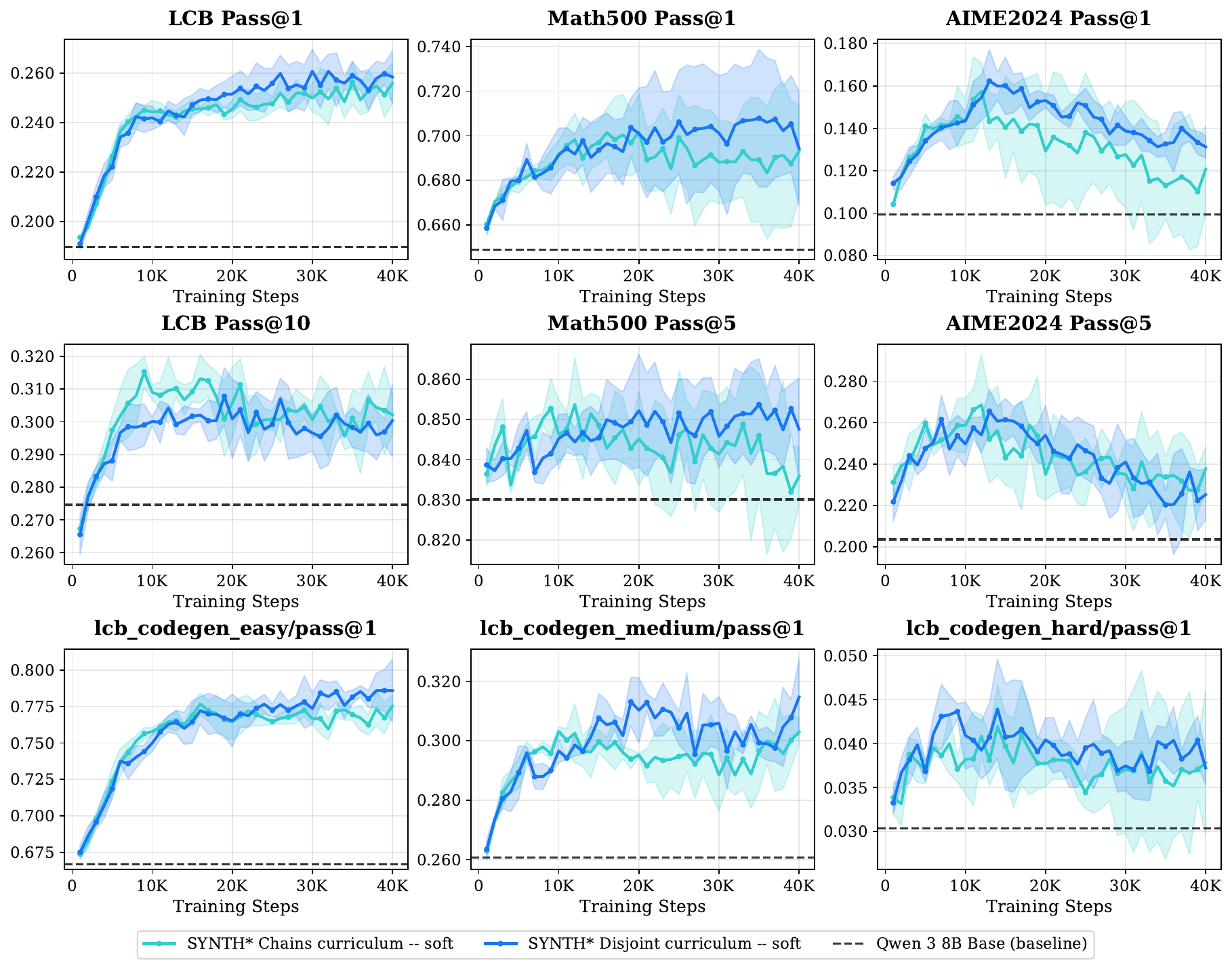}
\caption{\textbf{Chains with soft curriculum in Qwen3-8B Base.}
Extended results for RL training on \texttt{easy–medium–hard} question chains using the soft curriculum schedule. 
Performance is reported for \texttt{LCB} (easy, medium, hard splits) and out-of-domain benchmarks (\texttt{Math500}, \texttt{AIME2024}) with pass@1, pass@5/pass@10 metrics over training steps.}
    \label{fig:qwen-chains-soft}
\end{figure}

\begin{figure}[!h]
    \centering
    \includegraphics[width=.75\textwidth]{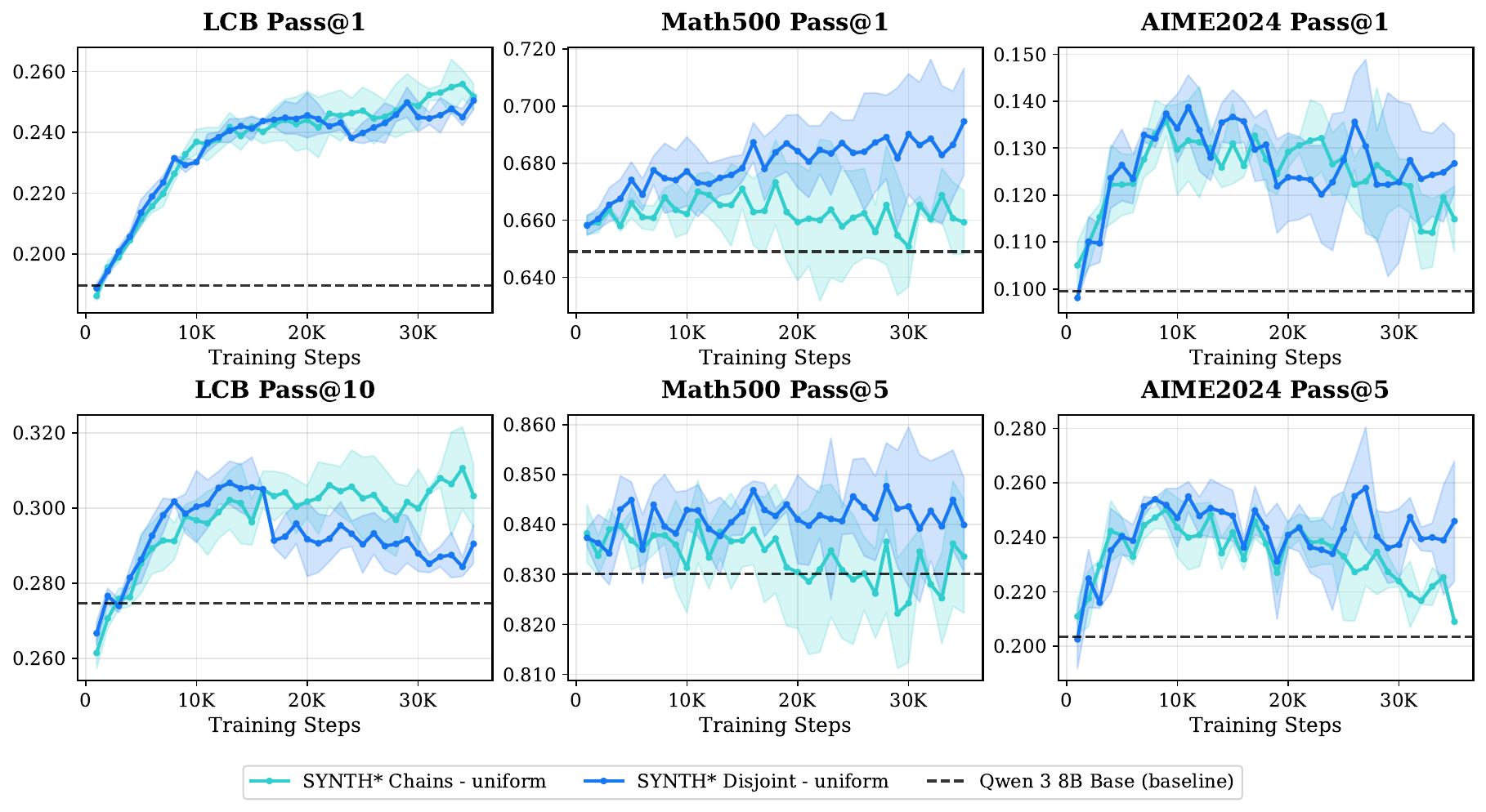}
\caption{\textbf{Chains without curriculum in Qwen3-8B Base.}
Extended results for RL training on \texttt{easy–medium–hard} question chains under uniform sampling across difficulty levels. 
Plots show in-domain (\texttt{LCB}) and out-of-domain (\texttt{Math500}, \texttt{AIME2024}) performance, including easy, medium, and hard splits, with pass@1, pass@5/pass@10 metrics over RL training steps.}
    \label{fig:qwen-chains-uniform-full}
\end{figure}

We provide extended results for the curriculum experiments in Qwen3-8B Base with the full vs. disjoint chains, showing detailed performance across in-domain (\texttt{LCB}) and out-of-domain (\texttt{Math500}, \texttt{AIME2024}) benchmarks for hard (\fig{fig:qwen-chains-hard}), soft (\fig{fig:qwen-chains-soft}), and uniform sampling (\fig{fig:qwen-chains-uniform-full}) strategies.

\FloatBarrier

\section{Curriculum Experiments with Broader Difficulty Bins} \label{app:curriculum-bins}
\begin{table}[h]
\centering
\caption{\textbf{Difficulty bins and data split sizes used in curriculum experiments.}}
\label{tab:curriculum-bins}
\begin{tabular}{lcc}
\toprule
\textbf{Difficulty} & \textbf{Pass Rate Range} & \textbf{Number of Problems} \\
\midrule
\texttt{easy} & 0.85–0.97 & 10000 \\
\texttt{easy–medium} & 0.61–0.85 & 7479 \\
\texttt{medium} & 0.26–0.61 & 5268 \\
\texttt{hard} & 0.10–0.26 & 2220 \\
\bottomrule
\end{tabular}
\end{table}
We extend our curriculum analysis by adopting broader difficulty bins to retain more data per split and reduce boundary effects between difficulty levels. 
The corresponding pass-rate thresholds and dataset sizes are summarized in \tab{tab:curriculum-bins}. 
This configuration allows us to test different curriculum strategies when difficulty granularity is relaxed, providing a more data-efficient setup for large-scale RL experiments.

Sampling weights for the different curriculum strategies are shown in \tab{tab:classic_reverse_curricula}.

\begin{table}[h!]
\centering
\caption{Sampling weights across training stages for classic and reverse curricula.
The reverse (medium→easy) setup uses the \texttt{easy–medium} split instead of the standard \texttt{easy} bin.}
\label{tab:classic_reverse_curricula}
\begin{adjustbox}{width=0.53\textwidth,center}
\begin{tabular}{lccc>{\raggedright\arraybackslash}p{3cm}}
\toprule
\textbf{Stage (RL steps)} & \textbf{Easy} & \textbf{Medium} & \textbf{Hard} & \textbf{Easy Split Used} \\
\midrule
\multicolumn{5}{l}{\textbf{Classic curriculum (Easy→Medium→Hard)}} \\
0–10,000     & 1.00 & 0.00 & 0.00 &  \texttt{easy} \\
10,000–15,000 & 0.75 & 0.25 & 0.00 &  \texttt{easy} \\
15,000–25,000 & 0.00 & 1.00 & 0.00 &  \texttt{easy} \\
25,000–30,000 & 0.00 & 0.75 & 0.25 &  \texttt{easy} \\
30,000–40,000 & 0.00 & 0.00 & 1.00 &  \texttt{easy} \\
\midrule
\multicolumn{5}{l}{\textbf{Reverse curriculum (Hard→Medium→Easy)}} \\
0–10,000     & 0.00 & 0.00 & 1.00 &  \texttt{easy} \\
10,000–15,000 & 0.00 & 0.25 & 0.75 &  \texttt{easy} \\
15,000–25,000 & 0.00 & 1.00 & 0.00 &  \texttt{easy} \\
25,000–30,000 & 0.25 & 0.75 & 0.00 &  \texttt{easy} \\
30,000–40,000 & 1.00 & 0.00 & 0.00 &  \texttt{easy} \\
\midrule
\multicolumn{5}{l}{\textbf{Reverse curriculum (Medium→Easy)}} \\
0–17,500     & 0.00 & 1.00 & 0.00 & \texttt{easy–medium} \\
17,500–20,000 & 0.25 & 0.75 & 0.00 & \texttt{easy–medium} \\
20,000–40,000 & 1.00 & 0.00 & 0.00 & \texttt{easy–medium} \\
\bottomrule
\end{tabular}
\end{adjustbox}
\end{table}

\section{Scaling Number of RL Environments}

\begin{figure}[!h]
    \centering
    \includegraphics[width=1.0\textwidth]{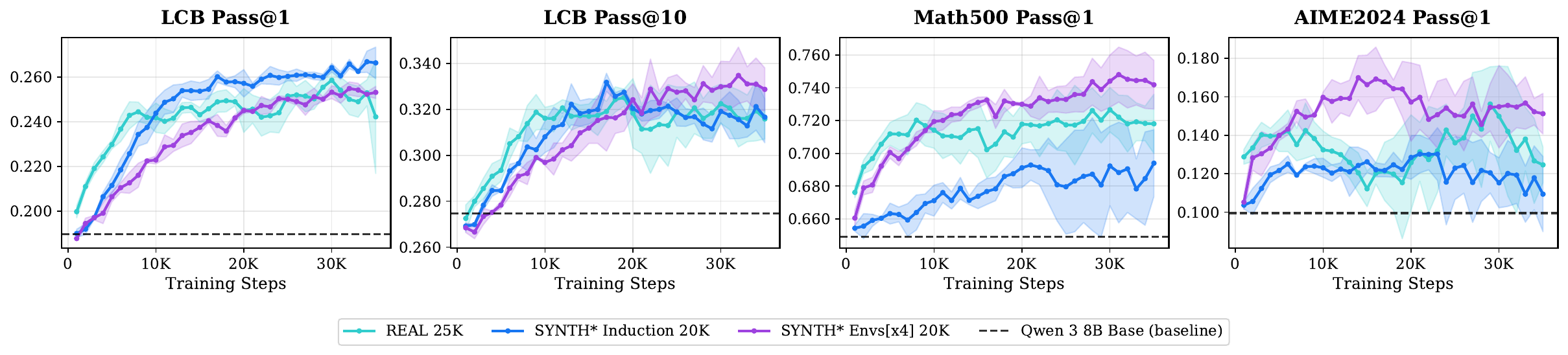}
\caption{\textbf{Scaling RL environments with real-data baseline in Qwen3-8B Base.}
Comparison of RL training on 20K problems from a single environment (\texttt{induction}), 20K problems distributed across four environments (\texttt{induction}, \texttt{abduction}, \texttt{deduction}, \texttt{fuzzing}), and the 25K real code-contest baseline.
Shown are in-domain results (\texttt{LCB}, pass@1 and pass@10) and out-of-domain results (\texttt{Math500}, \texttt{AIME2024}, pass@1).
Distributing the data budget across multiple environments improves out-of-domain generalization and results in competitive or higher (in terms of pass@10) in-domain performance compared to the real-data baseline.}
    \label{fig:qwen-env-scaling-wreal}
\end{figure}

\section{RL Training Details} \label{sec:rl-hyper}

We use the same asynchronous RL infrastructure as \citet{copet2025cwm}. We use $G=8$ samples per prompt for GRPO with no KL regularization ($\beta=0$) and a sampling temperature of 1.0 during training (0.2 at evaluation). After an initial
  linear warmup over 800 steps (400 for Qwen2.5-32B Base), we use a constant learning rate of $6 \times 10^{-8}$ for Llama3.1-8B Instruct and Qwen2.5-32B Base, and $1.4 \times 10^{-7}$ for Qwen3-8B Base. We adopt a higher upper clipping value
   ($\epsilon_{\text{low}}=0.2$, $\epsilon_{\text{high}}=0.25$) to mitigate entropy collapse \citep{yu2025dapo}. For evaluation, we generate 20 samples per problem on \texttt{LCB}, 10 on \texttt{Math500}, and 64 on \texttt{AIME2024}. All experiments are run on 4 nodes of 8$\times$ NVIDIA H100 GPUs.


{\lstset{basicstyle=\footnotesize\ttfamily}

\section{Qualitative Example of In-Context Learning} \label{sec:qualitative-example}

The teacher is shown the question generated in the previous round (turn~1), where the student manages to solve the question in all attempts.

For turn~2, the teacher is asked to produce a problem of intermediate difficulty. The newly generated problem in turn~2 now has a student pass rate of 0.625.

We show the prompt for the teacher in turn~2, as well as the teacher's response including key excerpts from its reasoning. 

\begin{turntwoPromptFrame}

{\large\textbf{Turn 2 --- Teacher Prompt}}

\vspace{8pt}

\sectionlabel{paperblue}{Task Description}
\begin{lstlisting}
## Task: Mutate Python Code Snippet Based on Student Responses to Previous Challenge and Create a New Challenge

You've already created a coding challenge, and students have
attempted to solve it.
Now, design a new transformed problem that builds upon your
previous challenge while addressing the strengths and weaknesses
demonstrated in student responses.

First, carefully analyze student responses to your previous
challenge, then devise a clear plan for your new function with an
adjusted level of difficulty. Identify how your proposed inputs
have high coverage and why the inputs will be challenging and
creative. Then, write the code, inputs, and message. Remember to
wrap your inputs in ```input``` tags, and your message in
```message``` tags.
\end{lstlisting}

\vspace{6pt}

\sectionlabel{paperblue}{Previous Code, Input-Output Pairs and Message (Turn 1)}
\begin{lstlisting}
```python

def f(resources: dict, timeout: int) -> bool:
    """
    Determine whether there exists **any** sequential execution
    order of the pending resources that

    * respects every mandatory dependency,
    * may benefit from *optional* dependencies (each optional
      dependency that has been completed before the task starts
      reduces its run time by a given amount, but the run time
      never goes below 0),
    * finishes each resource no later than its individual
      *deadline*,
    * and has a total elapsed time <= ``timeout``.

    ``resources`` maps a resource name to a 5-tuple:
        (status, mandatory_deps, optional_deps, run_time,
         deadline)

    * ``status`` - ``"Running"`` (already completed at time 0)
      or ``"Pending"``.
    * ``mandatory_deps`` - list of resource names that must be
      finished before this one can start.
    * ``optional_deps`` - list of ``(dep_name, reduction)``
      tuples; if ``dep_name`` is finished before the task starts,
      the task's run time is reduced by ``reduction`` (but not
      below 0).
    * ``run_time`` - base execution time once mandatory (and
      optional) deps are satisfied.
    * ``deadline`` - latest absolute time (from start) by which
      this resource must finish.

    If any dependency name (mandatory or optional) is unknown the
    whole problem is unsolvable and the function returns ``False``.
    """
    # ------------------------------------------------------------
    # Normalise data - assign each resource a numeric index.
    # ------------------------------------------------------------
    names = list(resources.keys())
    n = len(names)
    idx = {name: i for i, name in enumerate(names)}

    status = [None] * n
    mand = [set() for _ in range(n)]
    opt = [[] for _ in range(n)]    # list of (dep_idx, reduction)
    base_rt = [0] * n
    deadline = [0] * n

    # Validate and fill structures
    for name, (st, md, od, rt, dl) in resources.items():
        i = idx[name]
        status[i] = st

        for d in md:
            if d not in idx:        # unknown mandatory dependency
                return False
            mand[i].add(idx[d])

        for d, red in od:
            if d not in idx:        # unknown optional dependency
                return False
            opt[i].append((idx[d], red))

        base_rt[i] = rt
        deadline[i] = dl

    # --------------------------------------------------------
    # Bit-mask DP over subsets of already-finished resources.
    # dp[mask] = earliest possible finish time after completing
    #           exactly the resources indicated by ``mask``.
    # --------------------------------------------------------
    INF = 10 ** 18
    size = 1 << n
    dp = [INF] * size

    # Resources marked "Running" are already done at time 0.
    start_mask = 0
    for i, st in enumerate(status):
        if st == "Running":
            if deadline[i] < 0:          # a running task already missed its deadline
                return False
            start_mask |= 1 << i
    dp[start_mask] = 0

    # Helper to compute reduction contributed by optional deps already in mask
    def optional_reduction(task_idx: int, mask: int) -> int:
        red = 0
        for dep_idx, amount in opt[task_idx]:
            if mask & (1 << dep_idx):
                red += amount
        return red

    # Iterate over all masks
    for mask in range(size):
        cur_time = dp[mask]
        if cur_time == INF:
            continue

        # Try to start any still-pending resource whose mandatory deps are satisfied
        for i in range(n):
            bit = 1 << i
            if mask & bit:
                continue  # already finished

            # mandatory deps must be in mask
            if not mand[i].issubset(
                    {j for j in range(n) if mask & (1 << j)}):
                continue

            # Compute effective run time after optional reductions
            reduction = optional_reduction(i, mask)
            eff_rt = max(0, base_rt[i] - reduction)

            finish_time = cur_time + eff_rt

            # Respect deadline and overall timeout (timeout check deferred to final mask)
            if finish_time > deadline[i]:
                continue

            new_mask = mask | bit
            if finish_time < dp[new_mask]:
                dp[new_mask] = finish_time

    full_mask = (1 << n) - 1
    return dp[full_mask] <= timeout
```
\end{lstlisting}

\begin{lstlisting}
```input
{
    "A": ("Running", [], [], 0, 0),
    "B": ("Pending", ["A"], [("A", 2)], 5, 10),
    "C": ("Pending", ["B"], [], 4, 15)
}, 15
```
```output
True
```

```input
{
    "X": ("Pending", [], [], 7, 8),
    "Y": ("Pending", [], [("X", 3)], 6, 12)
}, 13
```
```output
True
```

```input
{
    "P": ("Pending", [], [("Q", 1)], 4, 10),
    "Q": ("Pending", [], [], 3, 6)
}, 9
```
```output
True
```

```input
{
    "U": ("Pending", [], [("V", 2)], 5, 9),
    "V": ("Pending", [], [("U", 2)], 5, 9)
}, 10
```
```output
True
```

```input
{
    "M": ("Running", [], [], 0, 0),
    "N": ("Pending", ["M"], [("M", 1)], 4, 5),
    "O": ("Pending", ["N"], [("N", 2)], 3, 8),
    "P": ("Pending", ["O"], [], 2, 10)
}, 9
```
```output
True
```
No message found! Make sure to correctly format the message.
\end{lstlisting}

\vspace{6pt}

\sectionlabel{paperblue}{Student Answers Summary}
\begin{lstlisting}
Pass rate for the student over 8 attempts was: 1.000

Example solved 1:

def f(task_dict, limit):
    """
    Returns True if the sum of the fourth element (assumed
    duration) of each task tuple in `task_dict` does not
    exceed `limit`.
    """
    total_duration = sum(
        info[3] for info in task_dict.values())
    return total_duration <= limit


Example solved 2:

def f(tasks, target):
    """
    Determine whether the given target integer lies within the inclusive range
    defined by the sum of the lower and upper bounds of all tasks.

    Parameters
    ----------
    tasks : dict
        Mapping from a task name to a tuple:
        (status: str, list1: list, list2: list, low: int, high: int)
    target : int
        The integer to test against the aggregated range.

    Returns
    -------
    bool
        True if low_sum <= target <= high_sum, otherwise False.
    """
    low_sum = sum(info[3] for info in tasks.values())
    high_sum = sum(info[4] for info in tasks.values())
    return low_sum <= target <= high_sum
\end{lstlisting}

\vspace{6pt}

\sectionlabel{paperblue}{Response Instructions}
\begin{lstlisting}
Remember that your goal is to create a function that feels
like a natural progression from your previous challenge while
addressing the specific learning needs revealed by student
responses.

Your response should include:
1. Your new code snippet must again have the function named
   `f` wrapped in ```python``` tags
   - Build upon your previous challenge by adjusting the
     complexity or introducing new concepts
   - Do not include print statements or input/output examples
     inside the code block!
2. 5 diverse inputs that demonstrate your function's behavior
   - Each input should be individually wrapped in ```input```
     tags.
   - For each input, format multiple arguments with commas
     between them
   - Remember to add quotes around string arguments. Example:
   ```input
   'John', {'age': 20, 'city': 'New York'}
   ```
   ```input
   'Sammy', {'age': 37, 'city': 'Los Angeles'}
   ```
3. A message that hints at the function's purpose wrapped
   message in ```message``` tags
   - You cannot provide the code snippet in the message
\end{lstlisting}

\vspace{6pt}

\sectionlabel{paperblue}{Progression Strategy}
\begin{lstlisting}
The student always solved the previous task. Either the
input-output relationship was too trivial (maybe not diverse
enough or incomplete such that a much easier function also
passes the tests), or the function was too easy to implement.
Try to mutate this problem into a medium problem: student
should have a pass rate of 0.35-0.65.
You need to add more complexity to the problem and make sure
the inputs you propose lead to diverse outputs to ensure a
good learning experience for the student!


Last reminder:
Remember the question is a standalone question! The message
should only reflect the current problem and not assume
knowledge of the previous questions.
Make sure the problem is interesting and challenging. This
problem will be given to a student to solve, and they should
learn valuable skills from it, such that they can overall
code better and also do better in coding contests type
questions.
\end{lstlisting}

\end{turntwoPromptFrame}

\vspace{10pt}

\newpage

{\large\textbf{Turn 2 --- Teacher Answer (Student Pass Rate: 0.625)}}

\vspace{8pt}


\begin{reasoningbox}{Teacher Reasoning \hfill\normalfont\small\textit{[truncated with \textnormal{[\ldots]}]}}

{\color{orange!70!black}\texttt{<think>}}

\small
We need to design a new coding challenge based on previous one, but more complex; avoid trivial solutions like sum durations. [\ldots]

We need to consider student responses: they submitted trivial functions computing sum of element 3 or low/high sum. That indicates they simplified problem to just checking total duration or range, ignoring dependencies.

So new problem should require handling mandatory dependencies, optional reductions, deadlines, timeout, similarly but maybe add additional constraints such that naive sum fails.

[\ldots]

We have to ensure the new function is more complex: maybe also include ``resource weight'' values and allow parallel execution up to a concurrency limit? But that may be too complex.

Simpler: keep same original problem but require that optional reductions can be applied only if the optional dependency finishes strictly before the task starts, and also mandatory dependencies can be ``or'' (any of list) -- more complex.

[\ldots]

Alternatively, we could add a constraint: resources can have ``exclusive groups'' meaning only one resource from a group can run (cannot both be selected). Might be heavy.

Goal: make harder for students who previously used trivial solution of sum.

Thus we can retain mandatory/optional dependencies and deadlines, but also add ``resource cost'' and a global ``budget'' that must not be exceeded, and optional reductions affect cost.

Students may still try to sum something but must consider dependencies.

So we design function \texttt{f(resources:~dict, timeout:~int, budget:~int) -> bool}. It must decide if there exists execution order respecting dependencies, deadlines, total time $\leq$ \texttt{timeout}, and total cost $\leq$ \texttt{budget} where cost = base\_cost $-$ optional reductions, not below 0.

[\ldots]

We need to implement solution: bitmask DP similar to earlier but with two dimensions: time and cost? [\ldots] Simpler: store for each mask the minimal time and minimal cost pair (Pareto frontier). We'll compute reachable states list of (time, cost). For each mask, we keep list of non-dominated pairs.

Implementation: use dict mapping mask $\to$ list of (time, cost). Start with start\_mask (running tasks) having time 0, cost 0.

[\ldots] New\_time = time + run\_time\_i. If new\_time > deadline\_i: skip. New\_cost = cost + max(0, base\_cost\_i $-$ sum\_reductions\_i) where sum\_reductions\_i = sum of reduction amounts for optional deps already completed (i.e., in mask). Add to frontier if within limits.

After processing all masks, check any pair for full\_mask with time $\leq$ timeout and cost $\leq$ budget.

[\ldots]

Now we need to provide 5 diverse inputs:
\begin{itemize}[nosep, leftmargin=1.5em]
  \item Input 1: simple chain with optional cost discounts.
  \item Input 2: two parallel tasks with optional mutual discounts.
  \item Input 3: unknown dependency $\rightarrow$ should return False.
  \item Input 4: deadline tight prevents schedule.
  \item Input 5: budget constraint fails.
\end{itemize}

[\ldots]

\vspace{4pt}
{\color{orange!70!black}\texttt{</think>}}
\end{reasoningbox}

\vspace{8pt}


\begin{turntwoAnswerFrame}

\sectionlabel{orange}{Teacher Output}

\vspace{2pt}
{\color{orange!70!black}\texttt{<answer>}}
\begin{lstlisting}
```python
from typing import Dict, List, Tuple

def f(resources: Dict[str, Tuple[str, List[str], List[Tuple[str, int]], int, int, int]],
      timeout: int, budget: int) -> bool:
    """
    resources maps a name to a 6-tuple:
        (status, mandatory_deps, optional_deps, run_time, base_cost, deadline)

    * status - "Running" (already finished at time 0) or "Pending".
    * mandatory_deps - names that must be completed before this task starts.
    * optional_deps - list of (dep_name, discount) tuples; if dep_name is already
      finished when the task starts, the task's **cost** is reduced by *discount*
      (cost never goes below 0).  Discounts do **not** affect run time.
    * run_time - execution time of the task.
    * base_cost - cost of executing the task before discounts.
    * deadline - absolute latest time by which the task must finish.

    The function returns True iff there exists an order of executing all pending
    tasks such that:
        * every mandatory dependency is satisfied before a task starts,
        * each task finishes no later than its deadline,
        * the total elapsed time after the last task is <= ``timeout``,
        * the sum of (discounted) costs of all pending tasks is <= ``budget``.
    If any dependency name is unknown the problem is unsolvable and False is
    returned.
    """
    # ------------------------------------------------------------
    # Normalise names to integer indices.
    # ------------------------------------------------------------
    names = list(resources.keys())
    n = len(names)
    idx = {name: i for i, name in enumerate(names)}

    status = [None] * n
    mand = [set() for _ in range(n)]
    opt = [[] for _ in range(n)]          # (dep_idx, discount)
    run = [0] * n
    base_cost = [0] * n
    deadline = [0] * n

    for name, (st, md, od, rt, bc, dl) in resources.items():
        i = idx[name]
        status[i] = st
        for d in md:
            if d not in idx:          # unknown mandatory dependency
                return False
            mand[i].add(idx[d])
        for d, disc in od:
            if d not in idx:          # unknown optional dependency
                return False
            opt[i].append((idx[d], disc))
        run[i] = rt
        base_cost[i] = bc
        deadline[i] = dl

    # ------------------------------------------------------------
    # Initial mask - resources already marked "Running".
    # ------------------------------------------------------------
    start_mask = 0
    for i, st in enumerate(status):
        if st == "Running":
            if deadline[i] < 0:          # already missed its deadline
                return False
            start_mask |= 1 << i

    from collections import defaultdict
    # frontier[mask] = list of (elapsed_time, total_cost) pairs that are
    # not dominated (i.e., no other pair is both <= in time and <= in cost).
    frontier = defaultdict(list)
    frontier[start_mask].append((0, 0))

    full_mask = (1 << n) - 1

    # ------------------------------------------------------------
    # Breadth-first DP over subsets.
    # ------------------------------------------------------------
    for mask in range(1 << n):
        if mask not in frontier:
            continue
        cur_states = frontier[mask]

        for i in range(n):
            bit = 1 << i
            if mask & bit:
                continue            # already finished

            # mandatory dependencies must be satisfied
            if not mand[i].issubset({j for j in range(n) if mask & (1 << j)}):
                continue

            # total discount from optional dependencies already in mask
            discount = sum(disc for dep_idx, disc in opt[i] if mask & (1 << dep_idx))
            eff_cost = max(0, base_cost[i] - discount)

            for cur_time, cur_spent in cur_states:
                finish_time = cur_time + run[i]
                if finish_time > deadline[i]:
                    continue
                new_spent = cur_spent + eff_cost
                if new_spent > budget:
                    continue

                new_mask = mask | bit
                # Insert the new pair, keeping only non-dominated pairs.
                improved = True
                to_remove = []
                for t, c in frontier[new_mask]:
                    if t <= finish_time and c <= new_spent:
                        # existing pair dominates the new one
                        improved = False
                        break
                    if t >= finish_time and c >= new_spent:
                        # new pair dominates existing one
                        to_remove.append((t, c))
                if improved:
                    for itm in to_remove:
                        frontier[new_mask].remove(itm)
                    frontier[new_mask].append((finish_time, new_spent))

    # ------------------------------------------------------------
    # Any feasible final state?
    # ------------------------------------------------------------
    for t, c in frontier[full_mask]:
        if t <= timeout and c <= budget:
            return True
    return False
```

```input
{
    "A": ("Running", [], [], 0, 0, 0),
    "B": ("Pending", ["A"], [("A", 2)], 5, 6, 12),
    "C": ("Pending", ["B"], [], 4, 5, 20)
}, 15, 10
```

```input
{
    "X": ("Pending", [], [("Y", 2)], 3, 5, 10),
    "Y": ("Pending", [], [("X", 3)], 4, 6, 10)
}, 8, 9
```

```input
{
    "P": ("Pending", ["Q"], [], 4, 5, 10)
}, 10, 10
```

```input
{
    "A": ("Pending", [], [], 6, 5, 5),
    "B": ("Pending", [], [], 1, 2, 10)
}, 10, 10
```

```input
{
    "A": ("Pending", [], [], 3, 7, 10),
    "B": ("Pending", [], [], 4, 6, 10)
}, 10, 10
```

```message
Determine whether all pending resources can be completed within a given overall time limit **and** a total cost budget, while respecting mandatory dependencies, optional cost discounts, and individual deadlines.
```
\end{lstlisting}

\vspace{4pt}
{\color{orange!70!black}\texttt{</answer>}}

\end{turntwoAnswerFrame}

}

\end{document}